%% file: main.tex
\documentclass[10pt]{article}
\usepackage[screen,lefttitle]{preprint}
\usepackage{placeins}
\usepackage{needspace}

\let\Phi\Phiit

\let\REQUIRE\Require
\let\ENSURE\Ensure
\let\STATE\State
\let\FOR\For
\let\ENDFOR\EndFor

\newcommand{\TO}{\textbf{to}}

\title{Heterogeneous Structured Pruning Under Resource Constraints for
Hardware-Efficient Large Language Models} %
\addauthor{Cameron Barker\thanks{Equal contribution.}}%
  {School of Informatics, University of Edinburgh, UK}{}
\addauthor{Andreas Paraskeva\samethanks}%
  {Leiden Institute of Advanced Computer Science, Leiden University, the Netherlands}{}
\addauthor{Jan N. van Rijn}%
  {Leiden Institute of Advanced Computer Science, Leiden University, the Netherlands}{}
\addauthor{Elliot J. Crowley}%
  {School of Engineering, University of Edinburgh, UK}{}
\addauthor{Henry Gouk}%
  {School of Informatics, University of Edinburgh, UK}{}
\preprintnotice{Preprint.}

\begin{document}
\newgeometry{left=2cm,right=1cm,top=1cm,bottom=1cm}
\maketitle

\input{sections/abstract}
\restoregeometry
\input{sections/introduction}
\input{sections/related-work}
\input{sections/structured-pruning}
\input{sections/hsp}
\input{sections/experiments}
\input{sections/conclusion}

\newpage

\input{sections/acknowledgements}
\printbibliography

\newpage
\appendix
\counterwithin{equation}{section}
\counterwithin{figure}{section}
\counterwithin{table}{section}

\input{sections/appendix/experimental-configuration}
\input{sections/appendix/procedure}
\input{sections/appendix/sparsity-support}
\input{sections/appendix/additional-results}
\input{sections/appendix/notation}

\end{document}

%% file: sections/abstract.tex
\begin{abstract}
  Training-free pruning removes weights from pretrained large language models to reduce their deployment cost.
  However, existing approaches typically consider only the total number of retained weights.
  Consequently, their pruning decisions do not account for how target hardware stores and processes those weights.
  These methods select a sparsity structure in advance and apply it uniformly to all projections,
  rather than accounting for how each structure trades pruning error against latency and memory across sublayers.
  We propose Heterogeneous Structured Pruning (HSP), a hardware-aware pruning framework that combines two components:
  (i)~a heterogeneous search space of dense, unstructured, $2{:}4$, block-sparse, and head / channel (H/C) pruning options
  and (ii)~structure-regularising constraints based on latency and memory models.
  Our contribution generalises existing importance-based methods:
  a global solver selects one option per sublayer to minimise proxy pruning loss under a resource budget,
  thereby assigning each projection a structure and density.
  We evaluate HSP on OPT and LLaMA models for an NVIDIA Jetson Orin Nano Super edge device, where it improves the quality--resource frontiers over training-free baselines.

  A project page and updated links for the related code is maintained at:\\
  \href{https://cameronbarker.me/publications/hsp}{\nolinkurl{cameronbarker.me/publications/hsp}}.
\end{abstract}

%% file: sections/introduction.tex
\begin{figure}[p]
  \centering
  \includegraphics[width=\textwidth]{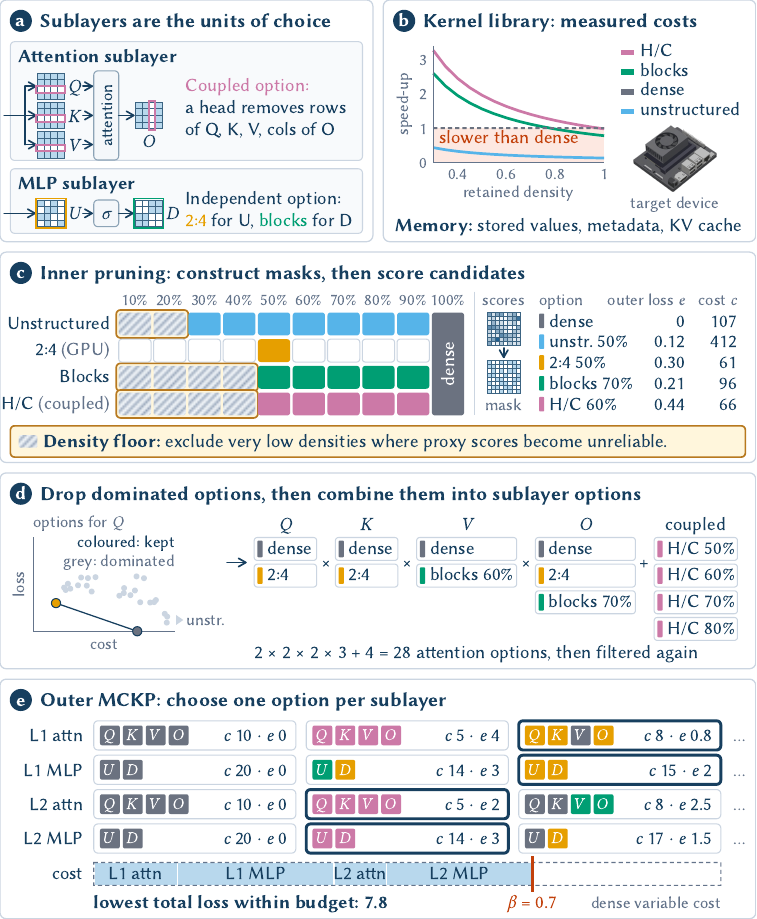}
  \caption{Overview of HSP.
    (a)~One option is selected per attention or MLP sublayer. H/C options couple projection masks, while independent options prune each projection separately.
    (b)~Kernel profiles give latency, and an encoding model gives memory.
    (c)~Density floors restrict candidates. Inner pruning constructs masks, outer scoring gives loss~$e$, and (b) gives cost~$c$.
    (d)~Dominance filtering reduces the candidate sets.
    (e)~MCKP chooses one option per sublayer. The bar shows variable cost against budget $B-C_0$, set here to a fraction $\beta=0.7$ of dense cost.
    Numbers and density grids are illustrative.}
  \label{fig:overview}
\end{figure}

\clearpage

\section{Introduction}

Rapid growth in the size of large language models (LLMs) has created a severe deployment bottleneck for resource-constrained edge devices.
Earlier work on efficient neural networks explored pruning, quantisation, and depthwise-separable convolutions,
including methods for deploying convolutional neural networks (CNNs) on mobile platforms \citep{LeCun1989optimal_brain_damage,han2015magnitudepruning,Wu2015QuantizedCN,han2016deepcompression,chollet2017xception}.
LLMs, however, pose a different pruning problem from CNNs \citep{zhu2024survey_model}.
CNNs reuse weights across spatial dimensions and are often compute-bound, whereas LLMs can be compute-bound or memory-bandwidth-bound depending on the inference phase.
These differences call for revisiting pruning structures chosen for CNNs and considering finer-grained sparsity within LLM projections.

Post-training pruning takes a trained model and removes weights to reduce the model's footprint while attempting to preserve accuracy.
One class of post-training pruning methods, which we refer to as retraining-free pruning, avoids the need for expensive recovery training and instead relies on lightweight heuristics to select weights for removal \citep{sun2024wanda,an2024flap}.
These methods typically use density (the proportion of remaining weights) as a proxy for model footprint, and they only consider a single sparsity structure across the entire model.
While unstructured pruning can achieve strong performance under a density budget, in this setting density is a poor proxy for deployment costs such as latency and memory footprint.
This has led to the development of structured pruning methods, which remove weights in a way that is more amenable to hardware acceleration.
However, these methods pick a single, coarse-grained sparsity structure a priori, which may degrade accuracy more than necessary if the target device supports a more fine-grained structure or the network contains critical layers that are sensitive to the chosen structure.

To address this gap, we propose Heterogeneous Structured Pruning (HSP),
a training-free framework that jointly optimises per-sublayer sparsity structures and densities under resource constraints.
Rather than relying on hardware-agnostic proxies such as parameter count, HSP uses profiled latency or analytically modelled memory to guide the pruning process.
We show that tailoring sparsity structures on a per-sublayer basis improves the trade-off between model performance and deployment efficiency.
Figure~\ref{fig:overview} summarises the framework.
Our main contributions are:
\begin{itemize}
  \item \textbf{Joint structure and density selection:}
    We formulate pruning as a bilevel problem: the inner problem selects masks, while the outer problem allocates structures and densities under a resource budget as a multiple-choice knapsack problem (MCKP).
  \item \textbf{Proxy-loss failure mode and density floors:}
    We identify a failure mode in which the importance-based proxy favours excessive pruning of a few projections or sublayers, and use structure-specific density floors to prevent these choices.
  \item \textbf{Sparse kernels and on-device evaluation:}
    We develop sparse kernels for the Jetson GPU and CPU and evaluate pruned models on both processors, measuring end-to-end latency and peak memory alongside model quality.
\end{itemize}

%% file: sections/related-work.tex
\section{Related Work}

{\bf Post-Training Pruning:}
Pruning may be integrated into training or applied to an already pretrained model.
Although post-training pruning (PTP) avoids pretraining a smaller model from scratch,
some PTP methods still use distillation or fine-tuning to recover quality after
components are removed \citep{muralidharan2024pruning_and_kd}.
Training-free methods differ in how they preserve quality without recovery training.
Wanda \citep{sun2024wanda} leaves retained weights unchanged,
FLAP \citep{an2024flap} adds bias compensation,
and SparseGPT \citep{frantar2023sparsegpt} updates retained weights through one-shot reconstruction.

{\bf Pruning Granularity:}
Pruning structures span a hierarchy of granularities.
At the coarsest level, methods such as ShortGPT and SLEB remove redundant transformer blocks \citep{men2025shortgpt,song2024sleb}.
Width pruning removes coupled channels or attention heads.
LLM-Pruner combines such removal with lightweight recovery tuning \citep{ma2023llm},
whereas SliceGPT reduces the embedding dimension by deleting rows and columns after orthogonal transformations \citep{ashkboos2024slicegpt}.
Finer-grained approaches remove weights individually,
in constrained $N{:}M$ groups \citep{mishra2021sparse},
or in block-sparse groups \citep{dalberto2024block}.
These impose different restrictions on the admissible masks
and have varying consequences for storage and execution.

{\bf Multi-Granularity Pruning:}
Recent methods combine several pruning granularities rather than committing to one throughout the model.
Sheared LLaMA jointly prunes layers, heads, and channels to a target architecture, followed by continued pretraining \citep{xia2024shearedllama}.
Cascaded Multi-Granularity Pruning removes layers, heads, and feed-forward channels in stages,
using low-rank recovery between stages \citep{wang2026cascadedmultigranularitypruningondevice}.
Both combine coarse-grained pruning with further training but do not consider finer-grained sparsity within projection matrices.

{\bf Hardware-Aware Pruning:}
Hardware-aware methods optimise directly against deployment cost rather than parameter count or FLOPs.
HALP uses a single channel-based structure in a latency-constrained global knapsack formulation for CNNs \citep{shen2026structural_pruning_lat_knapsack}.
Spatial downsampling, such as pooling, gives CNN layers different computation per weight which is exploited by HALP.
ZipLM targets runtime speedup by iteratively removing structured components with the least favourable loss-runtime trade-off,
and supports both one-shot and gradual compression \citep{kurtic2023ziplm}.
MDP jointly prunes channels, heads, and embeddings for vision workloads through a
mixed-integer nonlinear program constrained by predicted latency \citep{sun2025mdpmultidimensionalvisionmodel}.
Related hardware-aware neural architecture search (HW-NAS) methods compare candidate architectures using measured or estimated device costs \citep{li2021hwnasbench}.

HSP jointly selects pruning structures and densities across LLM sublayers (attention or MLP) without further training,
considering both coarse-grained structures and finer-grained sparsity within projection matrices.
Its bilevel formulation separates mask construction by existing pruning procedures from global structure and density selection.
HSP uses profiled latency or analytically modelled memory rather than a density proxy.
To our knowledge, HSP is the first to combine these granularities in training-free LLM pruning under a modelled latency or memory budget.

%% file: sections/structured-pruning.tex
\section{Pruning with Structured Sparsity}
\label{sec:structured_pruning}

\begin{figure}[!tp]
  \centering
  \captionsetup[subfigure]{font={normalsize,sf,bf},justification=centering,
    singlelinecheck=false,position=top,skip=3pt,hypcap=true}
  \begin{subfigure}{\linewidth}
    \centering
    \caption{Weight memory layouts at equal density}
    \label{fig:structure_memory_layouts}
    \includegraphics[width=\linewidth]{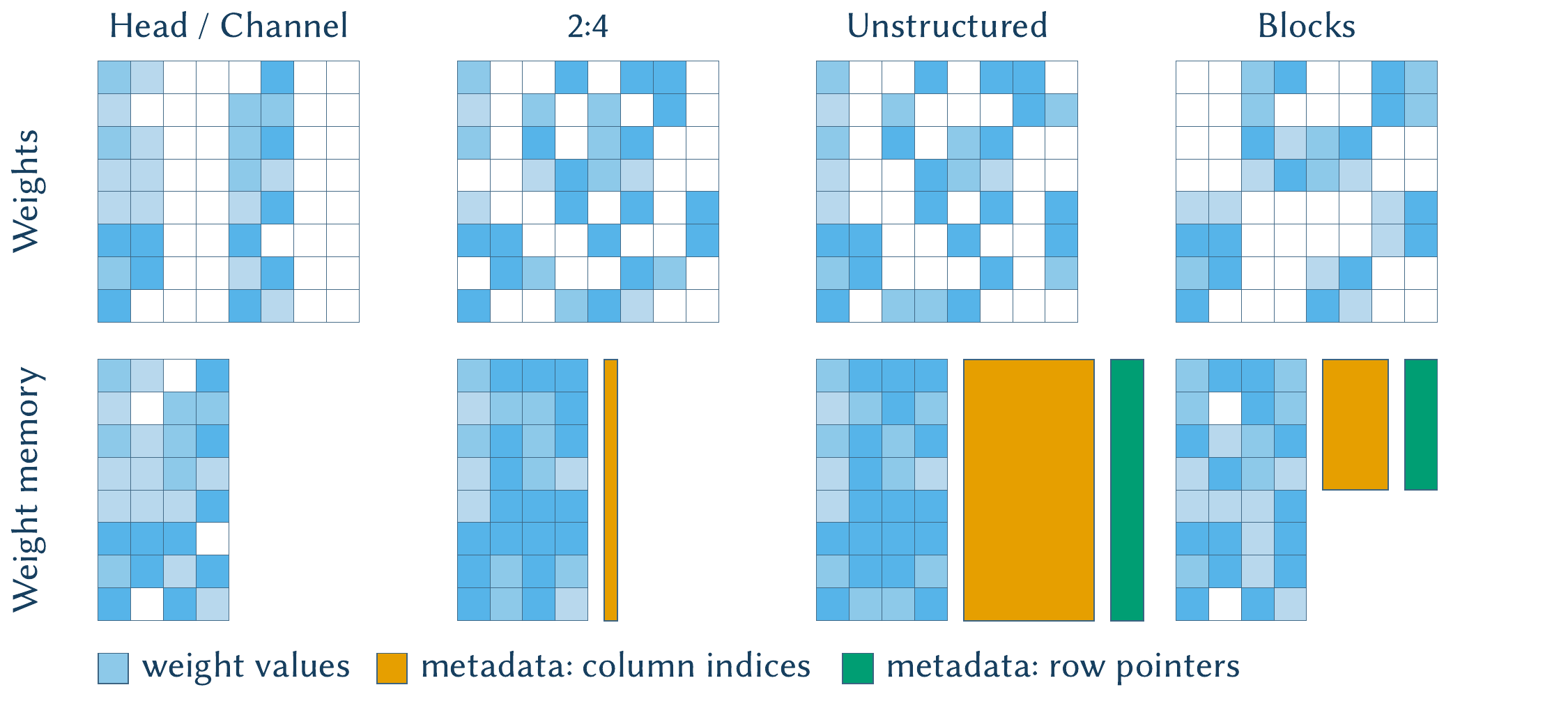}
  \end{subfigure}
  \par\medskip
  \begin{subfigure}{\linewidth}
    \centering
    \caption{OPT-2.7B \textperiodcentered{} query-projection weight memory}
    \label{fig:projection_memory}
    \includegraphics[width=\linewidth]{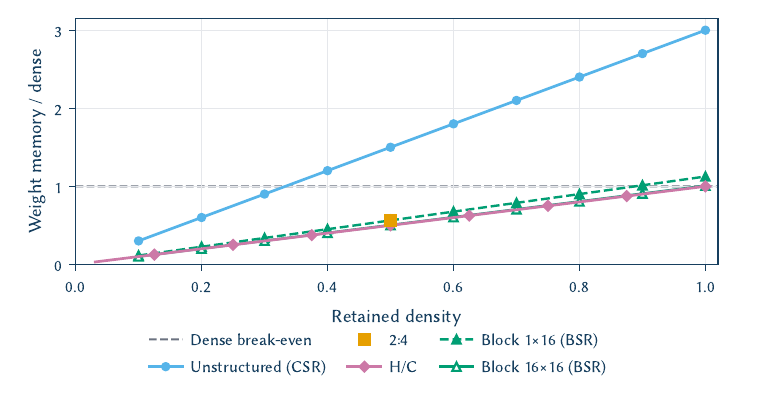}
  \end{subfigure}
  \caption{Structure-dependent weight memory.
    (\subref*{fig:structure_memory_layouts})~Equal-density masks need different amounts of metadata. Orange denotes column indices and green denotes row pointers.
    (\subref*{fig:projection_memory})~Analytical weight memory for the $2560\times2560$ OPT-2.7B query projection, relative to dense FP16 weights.
    Compressed sparse row (CSR) and block sparse row (BSR) formats use 32-bit indices and row pointers.
    The $2{:}4$ format uses packed metadata and has fixed density $0.5$.
    H/C pruning stores the narrowed matrix densely.
    Values below the dashed line save weight memory.
    Costs include weights and metadata only, excluding biases, cache, and runtime allocations.}
  \label{fig:structure_memory}
\end{figure}

Removing weights reduces cost only when the target hardware, software kernel,
and storage format can exploit the resulting sparsity.
Supported structures differ across devices:
coarse pruning maps naturally to smaller dense operations,
whereas fine-grained sparsity may require specialised hardware primitives
(e.g., NVIDIA's $2{:}4$ sparsity)
and otherwise fall back to slower general-purpose paths.
Software is often the limiting factor,
since hardware support provides speed-up only when exposed through efficient kernels for the relevant matrix shapes.
Weight memory is similarly structure-dependent:
irregular formats require more positional metadata than block-sparse formats or the compact dense matrices produced by width pruning,
so equal retained density need not imply equal memory footprint (Figure~\ref{fig:structure_memory}).
\subsection{Pruning Under a Density Budget}

Let $\Pi$ index the prunable linear projections of the model,
$\mathbf{W}=\{\mathbf{W}^{(\pi)}\}_{\pi\in\Pi}$ their pretrained weights,
and $\mathbf{M}=\{\mathbf{M}^{(\pi)}\}_{\pi\in\Pi}$ corresponding binary masks of the same shapes.
Indices $i$ and $j$ identify output rows and input columns.
$M^{(\pi)}_{i,j}=1$ retains weight $W^{(\pi)}_{i,j}$ and $0$ removes it.
Let $\mathcal{M}^{(\pi)}$ be the set of all binary masks for projection $\pi$,
and $\mathcal{M}=\prod_{\pi\in\Pi}\mathcal{M}^{(\pi)}$ their joint mask space.
The retained density is
$D(\mathbf{M})=\sum_{\pi\in\Pi}\lVert\mathbf{M}^{(\pi)}\rVert_0/\allowbreak\sum_{\pi\in\Pi}\operatorname{numel}(\mathbf{W}^{(\pi)})$,
where $\lVert\cdot\rVert_0$ counts nonzero entries and $\operatorname{numel}(\cdot)$ counts all entries.
We retain the convention used in LLM pruning work such as Wanda~\citep{sun2024wanda}:
density counts only the prunable projection weights.
Untouched parameters, such as embedding weights, still contribute to the model's total parameter count.

Let $\mathcal{L}(\cdot;\mathcal{D})$ denote the task loss on a dataset $\mathcal{D}$.
For an aggregate density target $\rho\in[0,1]$, the ideal pruning objective is
\begin{equation}
  \mathbf{M}_{\rho}^{\star}
  =
  \underset{\mathbf{M}\in\mathcal{M}}{\arg\min}\;
  \mathcal{L}(\mathbf{W}\odot\mathbf{M};\mathcal{D})
  \quad\mathrm{s.t.}\quad
  D(\mathbf{M})\leq\rho ,
  \label{eq:density_objective}
\end{equation}
where $\mathbf{M}_{\rho}^{\star}$ is an optimal mask collection and $\odot$ denotes elementwise masking of each projection.
The loss is evaluated on the resulting model, with parameters outside the pruning scope held fixed.
In practice, mask selection uses the importance proxy below.

{\bf Importance Scores:}
Evaluating $\mathcal{L}$ for every candidate mask is intractable.
Pruning methods use inexpensive heuristic scores as proxies for each weight's contribution to model performance.
These scores are computed from the pretrained weights and,
optionally, activations collected on a small training-data calibration set $\mathcal{D}_{\mathrm{cal}}$.
For projection $\pi$, let $\mathbf{S}^{(\pi)}$ be its matrix of weight-importance scores, with the same shape as $\mathbf{W}^{(\pi)}$.
Its input-activation matrix $\mathbf{X}^{(\pi)}$ has one row per calibration token and one column per input feature.
Magnitude pruning \citep{han2015magnitudepruning} uses the absolute weight,
whereas Wanda scales this magnitude by the norm of the
corresponding input feature,
\begin{equation}
  S^{(\pi)\mathrm{Mag}}_{i,j}
  =
  \lvert W^{(\pi)}_{i,j}\rvert,
  \qquad
  S^{(\pi)\mathrm{Wanda}}_{i,j}
  =
  \lvert W^{(\pi)}_{i,j}\rvert
  \cdot
  \lVert \mathbf{X}^{(\pi)}_{:,j}\rVert_2 ,
  \label{eq:importance_scores}
\end{equation}
where $\mathbf{X}^{(\pi)}_{:,j}$ contains all calibration-token activations of input feature $j$
and $\lVert\cdot\rVert_2$ is the Euclidean norm.
For either score matrix, the removed-importance proxy loss $\ell$ is
\begin{equation}
  \ell\bigl(\mathbf{M}^{(\pi)}, \mathbf{S}^{(\pi)}\bigr)
  =
  \sum_{i,j} \bigl(1-M^{(\pi)}_{i,j}\bigr)S^{(\pi)}_{i,j} ,
  \label{eq:proxy_loss}
\end{equation}
The aggregate proxy loss is the sum over projections,
$\widetilde{\mathcal{L}}(\mathbf{M})=\sum_{\pi\in\Pi}\ell(\mathbf{M}^{(\pi)},\mathbf{S}^{(\pi)})$,
which replaces $\mathcal{L}$ in Equation~\ref{eq:density_objective}.

{\bf Pruning Structures:}
A sparsity structure $s$ specifies which patterns of retained weights are permitted.
Together with a projection-level retained density $d\in[0,1]$, it defines the set
$\mathcal{M}^{(\pi)}(s,d)\subseteq\mathcal{M}^{(\pi)}$ of admissible masks.
Unstructured masks permit individual weight removal. Wanda additionally fixes the retained count in each output row.
Block-sparse masks retain or remove entire rectangular blocks,
whereas $2{:}4$ masks retain two weights in every window of four consecutive input positions, fixing $d=0.5$.
These restrictions allow compatible storage formats and kernels to exploit sparsity.
We seek a mask for every prunable projection $\pi\in\Pi$.
Given fixed importance scores, structure and density, these masks can be selected independently
when the admissibility constraints involve only the respective projection:
\begin{equation}
  \mathbf{M}^{\star(\pi)}(s,d)
  =
  \underset{\mathbf{M}^{(\pi)}\in\mathcal{M}^{(\pi)}(s,d)}{\arg\min}\;
  \ell\bigl(\mathbf{M}^{(\pi)},\mathbf{S}^{(\pi)}\bigr) .
  \label{eq:projection_selection}
\end{equation}
Wanda uses this per-projection mask selection with a uniform structure and density across projections.

{\bf Coupled Structures:}
Head / Channel (H/C) structures constrain the masks of multiple projections jointly.
We treat the attention and multilayer perceptron (MLP) components of each decoder layer as separate \emph{sublayers}.
Each sublayer, indexed by $l$, has a set of projections $\Pi_l\subseteq\Pi$ that share intermediate activations.
For standard multi-head attention, these are the query, key, value, and output projections $\{Q,K,V,O\}$.
For a two-projection MLP, they are the up and down projections $\{U,D\}$.
Let $d_h$ be the dimension of each attention head.
Removing a head removes $d_h$ corresponding rows from $Q$, $K$, and $V$ and columns from $O$.
For MLPs, removing a hidden channel implies a mask on $U$ that removes a row and a corresponding column of $D$.
The affected projections therefore cannot choose their masks independently.
We call the projection that consumes the shared intermediate activations the \emph{consumer}, denoted $\pi_{\mathrm{c}}$: $O$ for attention and $D$ for the MLP.
The upstream projections that produce these features are the \emph{producers}: $Q$, $K$, and $V$ for attention, and $U$ for the two-projection MLP.
We describe extensions for gated MLPs and grouped-query attention in Section~\ref{app:coupled_structures}.

\begin{figure}[!t]
  \centering
  \includegraphics[width=\textwidth]{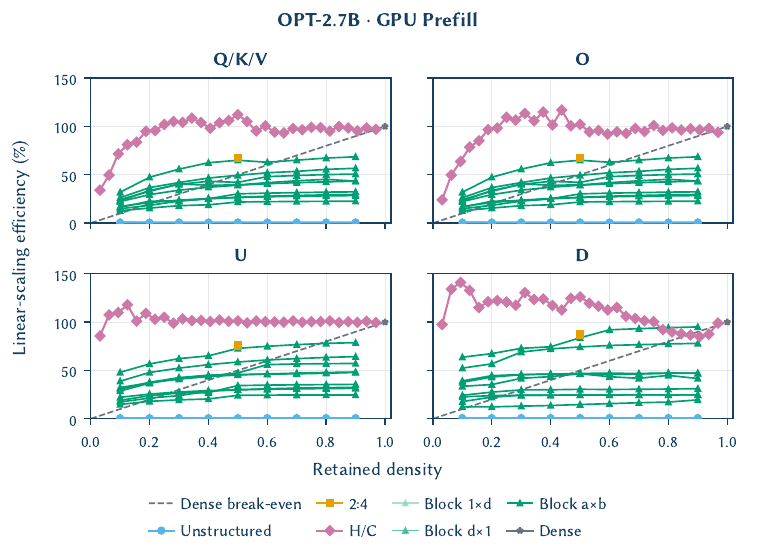}
  \caption{Measured GPU-prefill kernel performance for OPT-2.7B on the Jetson Orin Nano Super.
    Panels show the Q/K/V, attention output (O), MLP up (U), and MLP down (D) projections.
    Linear-scaling efficiency is $E_{\mathrm{lin}}(d)=d\,T_{\mathrm{dense}}/T_{\mathrm{sparse}}$.
    Here $d$ is retained density, and $T_{\mathrm{dense}}$ and $T_{\mathrm{sparse}}$ are measured latencies of the dense and sparse kernels for the same projection and workload.
    Points above the dashed $100d\%$ line are faster than dense execution.
    A value of $100\%$ denotes ideal inverse-density scaling.
  Each rectangular block shape is plotted separately and grouped into a block family in the legend.}
  \label{fig:kernel_efficiency_qoud}
\end{figure}

\subsection{Density Is Not a Deployment Cost}
\label{sec:density_deployment_cost}

Imposing a structure restricts the admissible mask space.
At fixed density, the space of all binary masks contains every structured mask,
so its minimum proxy loss cannot exceed theirs.
Consequently, an objective constrained only by $D(\mathbf{M})\leq\rho$ provides no incentive to select a deployable structure,
even though deployment cost depends jointly on density, structure, hardware, and the available kernel or storage implementation.

Figure~\ref{fig:projection_memory} quantifies the weight memory differences using the encoding formulas in Equation~\ref{eq:app_projection_memory}.
At $50\%$ retained density, CSR needs approximately $1.50\times$ the dense weight memory,
compared with $0.5625\times$ for $2{:}4$ and $0.50\times$ for H/C pruning.
The benefit depends on how the retained weights and their positions are encoded, not just how many weights remain.

We develop sparse kernels for the Jetson GPU and CPU and profile them alongside available implementations.
Figure~\ref{fig:kernel_efficiency_qoud} shows GPU-prefill measurements for OPT-2.7B on the Jetson Orin Nano Super.
At equal density, kernel efficiency varies across structures and projections.
The most fine-grained sparse kernels remain slower than dense execution despite removing most weights.
Together with the weight memory differences in Figure~\ref{fig:structure_memory},
this motivates jointly selecting structures and densities under a resource budget.
Kernel details and further measurements appear in Sections~\ref{app:kernels}--\ref{app:measurement_protocol}.

%% file: sections/hsp.tex
\section{Heterogeneous Structured Pruning}

Heterogeneous Structured Pruning (HSP) is a training-free framework for selecting sparsity structures and densities under a deployment-resource budget.
Its bilevel formulation separates two decisions: the inner problem constructs a pruned mask for each candidate option, and the outer problem chooses one option per sublayer.
HSP reuses the importance scores and masks selected by standard pruning methods for each candidate option, then compares these candidates globally by quality and hardware cost (Figure~\ref{fig:overview}).

\subsection{Resource-Constrained Bilevel Pruning}
\label{sec:hsp_bilevel}
For each of the $L$ attention and MLP sublayers, indexed by $l$, let $\Phi^{(l)}$ be a finite set of pruning options.
An option $\phi^{(l)}$ specifies the structure and density of every projection in $\Pi_l$, including any coupling between their masks.
Let $\mathbf{M}^{(l)}=(\mathbf{M}^{(\pi)})_{\pi\in\Pi_l}$ collect these masks, and let $\mathcal{M}^{(l)}(\phi^{(l)})$ be their admissible set.
The inner loss $\ell_{\mathrm{in}}^{(l)}$ determines which admissible mask to construct.
The outer loss $\ell_{\mathrm{out}}^{(l)}$ evaluates the resulting mask, while $C_l(\phi^{(l)})$ gives the option's resource cost.

A model-wide configuration is $\boldsymbol{\phi}=(\phi^{(l)})_{l=1}^{L}\in\Phi$, where $\Phi=\prod_l\Phi^{(l)}$.
With absolute resource budget $B$ and fixed cost $C_0$ outside the option space, HSP solves
\begin{equation}
  \begin{aligned}
    \boldsymbol{\phi}^{\star}
    &=\underbrace{
      \underset{\boldsymbol{\phi}\in\Phi}{\arg\min}\;
      \sum_{l=1}^{L}\ell_{\mathrm{out}}^{(l)}
      \!\left(\mathbf{M}^{\star(l)}(\phi^{(l)});\phi^{(l)}\right)
    }_{\text{Outer problem: MCKP selection}}\\[4pt]
    &\mathbf{M}^{\star(l)}(\phi^{(l)})
    =\underbrace{
      \underset{\mathbf{M}^{(l)}\in\mathcal{M}^{(l)}(\phi^{(l)})}{\arg\min}\;
      \ell_{\mathrm{in}}^{(l)}(\mathbf{M}^{(l)};\phi^{(l)})
    }_{\text{Inner problem: standard pruning}},\\
    &\mathrm{s.t.}\quad C_0+\sum_{l=1}^{L}C_l(\phi^{(l)})\leq B.
  \end{aligned}
  \label{eq:hsp_objective}
\end{equation}
Here $\mathbf{M}^{\star(l)}(\phi^{(l)})$ is the mask collection selected by the inner problem for option $\phi^{(l)}$ and is solved for each candidate before the outer allocation.
The outer problem selects one option per sublayer from these fixed loss--cost pairs under the global budget, forming a multiple-choice knapsack problem (MCKP).
Before allocation, we discard Pareto-dominated options first within each projection (linear layer), before combining independent projection candidates, and then within each attention or MLP sublayer, including coupled candidates.
Removing these options preserves the optimal loss under the resource budget~\citep{sinha1979mckp}.
We solve the resulting allocation approximately with a Pareto-frontier dynamic program~\citep{kellerer2004knapsack}.
The solver details are in Section~\ref{app:mckp_solver}.

\subsection{Candidate Masks and Density Floors}
\label{sec:option_sets}

{\bf Independent and Coupled Candidates:}
We consider two types of candidates: independent candidates for pruning within individual projections, and coupled candidates for pruning shared heads or channels across projections.
An independent option assigns each projection a dense, unstructured, $2{:}4$, or rectangular-block structure and a retained density.
Its masks are constructed separately using Equation~\ref{eq:projection_selection}.
A coupled H/C option instead chooses shared attention heads or MLP channels and applies the coupled masks defined in Section~\ref{sec:structured_pruning}.
Following the fixed-density WandaSP construction described by \citet{an2024flap}, we select these shared units using the consumer projection's scores: $O$ for attention and $D$ for the MLP.
The masks on the producer projections follow from this selection.
The precise option sets and mask construction are described in Section~\ref{app:option_construction}.

{\bf Scoring the Resulting Masks:}
By default we use Wanda scores for both the inner and outer losses however it is possible for each to use different importance metrics.
We explore using magnitude scores in Section~\ref{app:importance_ablation}.
For the outer loss, each projection's scores are divided by their mean, with an all-zero score matrix left unchanged.
This preserves the within-projection ranking and equalises the mean score per weight. Larger projections still contribute more total importance.
Write the normalised outer scores as $\widehat{\mathbf{S}}^{(\pi)}$.
For both independent and coupled candidates, the outer loss evaluates every affected projection:
\begin{equation}
  \ell_{\mathrm{out}}^{(l)}(\mathbf{M}^{(l)};\phi^{(l)})
  =w^{(l)}(\phi^{(l)})
  \sum_{\pi\in\Pi_l}
  \ell\!\left(\mathbf{M}^{(\pi)},
                  \widehat{\mathbf{S}}^{(\pi)}\right),
  \label{eq:outer_loss}
\end{equation}
where $\mathbf{M}^{(\pi)}$ is projection $\pi$'s mask within $\mathbf{M}^{(l)}$ and $w^{(l)}(\phi^{(l)})>0$ is an option weight.
Equation~\ref{eq:hsp_objective} evaluates this loss at the inner solution $\mathbf{M}^{\star(l)}(\phi^{(l)})$.
Thus, coupled candidates are constructed using the consumer alone but scored using the entire sublayer.
Section~\ref{app:option_construction} gives the normalisation formula, and Section~\ref{app:experimental_configuration} gives the experimental option weights.

{\bf Density Floors:}
Without a lower limit on candidate densities, minimising the model-wide proxy loss concentrates severe pruning in a few projections or sublayers, and language-model quality collapses.
At these extreme densities, a low proxy loss does not indicate a good pruned model.
We prevent this collapse with a density floor: each structure $s$ admits only candidate densities $d\geq d_{\min}(s)$, where $d_{\min}(s)>0$.
Above the floor, the allocation can still choose heterogeneous densities.
The density grids and floor settings are given in Section~\ref{app:experimental_configuration}.

\subsection{Resource Models and Budget Scope}
\label{sec:hsp_resources}
For either resource objective, we sum component costs within each sublayer and then across sublayers:
\begin{equation}
  C(\boldsymbol{\phi})=C_0+\sum_{l=1}^{L}
  \underbrace{\left[
    \sum_{\pi\in\Pi_l}C_{\pi}(\phi^{(l)})
    +C_{\mathrm{aux}}^{(l)}(\phi^{(l)})
  \right]}_{C_l(\phi^{(l)})}.
  \label{eq:cost_separable}
\end{equation}
Here $C_{\pi}$ is the cost of an individual linear projection, while $C_{\mathrm{aux}}^{(l)}$ collects significant auxiliary costs within the sublayer beyond its linear projections such as the attention operation and key--value cache which varies under head pruning.

For latency and memory, $B$ is a fixed budget in units of time or bytes, respectively.
Since $C_0$ is fixed, the solver allocates the remaining budget $B-C_0$ across sublayers.

{\bf Latency:}
For each target device and inference phase, we profile every unique combination of projection dimensions, sparsity structure, and retained density in the option space, along with the induced attention configurations.
Projections with matching configurations reuse the same measured latency, so profiling need not be repeated for every layer or budget.
Here $C_0$ covers the profiled dense output head and any input/output embedding-width projections outside the pruning scope.
These profiles estimate operator latency.
Separately, the on-device results tables report the actual real-world end-to-end execution time.

{\bf Memory:}
We count FP16 projection weights, sparse metadata, retained biases, and the peak key--value cache during the workload.
The fixed term counts untouched parameters such as the embedding and output head.
Projection biases belong to the sublayer costs because H/C options can narrow them.
This analytical model excludes temporary activations, allocator overhead, and kernel workspaces, and is distinct from the measured peak memory in the on-device tables.

The encoding formulas and profiling protocol are in Section~\ref{app:sparsity_hardware}.

\subsection{From Solution to Pruned Model}
The selected options determine the masks to apply, with no further search.
Options are implemented using the same memory format and related kernel that the profiler picked for their structure and density.
Head and channel pruning is implemented by converting layers into narrower dense projections and thus use the same kernels as the original model.
Shared activations or attention heads and associated cached key--value pairs that are pruned are thus removed without needing specialised kernels.
Following FLAP~\citep{an2024flap}, we compensate for removed input contributions by adjusting the consumer bias for the coarser options for models that already have biases.

Section~\ref{app:solver_and_realisation} gives the algorithm and realisation details.

%% file: sections/experiments.tex
\section{Experiments}

\begin{figure*}[t]
  \centering
  {\small\bfseries Mixed prefill and decode, GPU\par}
  \includegraphics[width=\textwidth,trim=0 0 0 5bp,clip]{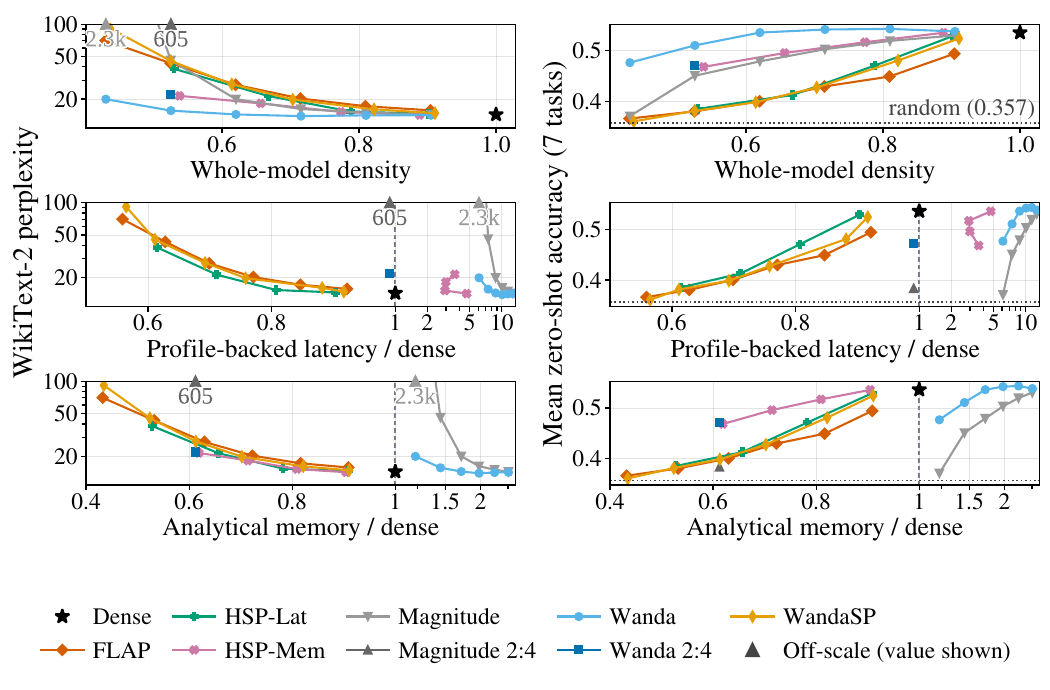}
  \caption{WikiText-2 perplexity and mean seven-task zero-shot accuracy vs. whole-model density,
  profile-backed latency, and analytical memory (normalised to dense) for
  OPT-2.7B on the Jetson Orin Nano Super GPU under the mixed workload (1024-token prefill, 128-token decode).
  Resource axes are linear below dense cost and logarithmic above.
  Triangles ($\blacktriangle$) mark perplexities above the plot limit of $100$, with numeric annotations giving their values.
  The dashed rule marks random-guess accuracy (0.357).}
  \label{fig:frontier_ppl_2p7b_gpu_mixed}
\end{figure*}

We evaluate two variants of our method: HSP-Lat, which selects pruning configurations under a profiled-latency budget, and HSP-Mem, which selects them under an analytical-memory budget.
Our main experiments use the OPT model family \citep{zhang2022opt}.
We compare against the following baselines: the dense model, unstructured Magnitude and Wanda, their $2{:}4$ variants, FLAP, and WandaSP.
We evaluate HSP-Lat and HSP-Mem against the baselines using perplexity on the WikiText-2 test set~\citep{merity2017pointer}
and zero-shot accuracy on BoolQ, PIQA, HellaSwag, WinoGrande, ARC-e, ARC-c, and OpenBookQA \citep{clark2019boolq,Bisk2020,zellers2019hellaswag,sakaguchi2019winogrande,allenaiarc,OpenBookQA2018}.
Perplexity measures retained next-token prediction quality, while zero-shot accuracy assesses performance on downstream language-understanding tasks.
We also measure peak memory and end-to-end latency for selected HSP candidates and baselines on the target device.
We provide evaluation-metric definitions, randomness controls, and hardware setup in Section~\ref{app:experimental_configuration}, and profiling and kernel details in Section~\ref{app:sparsity_hardware}.
Section~\ref{app:importance_ablation} ablates the importance score and coupled H/C pruning, while Section~\ref{app:calibration} examines calibration sample-count sensitivity.
Section~\ref{app:scaling} reports additional results for smaller OPT models, LLaMA 7B, and different workloads.
Wall-clock pruning and profiling costs are reported in Section~\ref{app:computational_cost}.

\subsection{Quality-Resource Frontiers}
\label{sec:quality_resource_pareto}
\begin{figure*}[!t]
  \centering
  {\small\bfseries Mixed prefill and decode, CPU\par}
  \includegraphics[width=\textwidth,trim=0 0 0 5bp,clip]{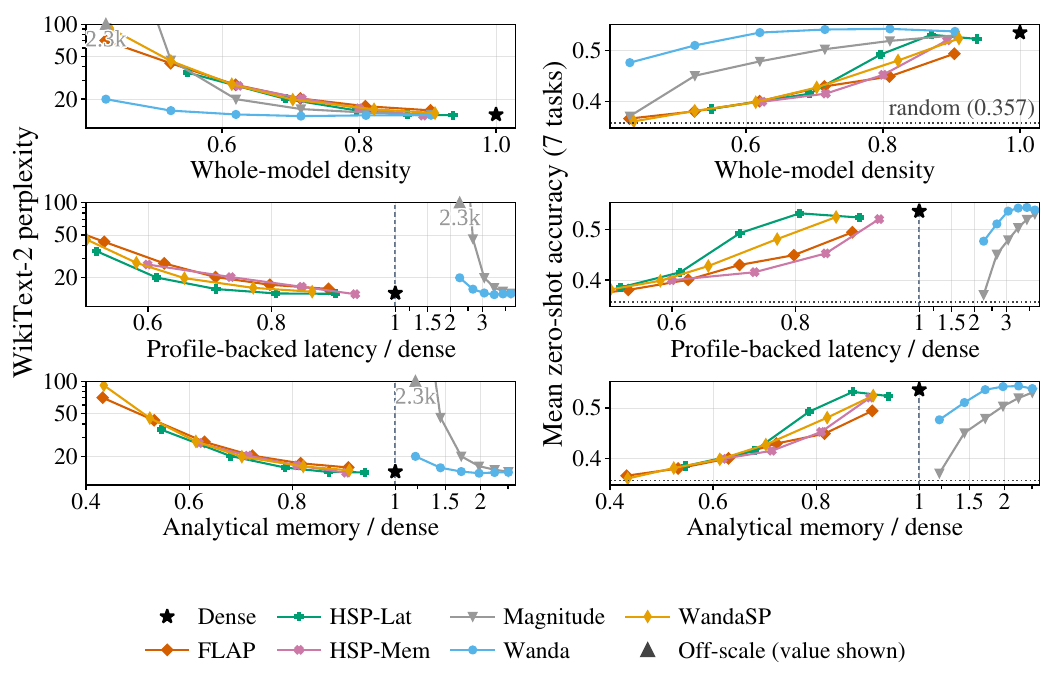}
  \caption{
    WikiText-2 perplexity (left) and mean seven-task zero-shot accuracy (right) vs. density, latency, and memory for OPT-2.7B on the Jetson CPU.
    Plotting conventions follow Figure~\ref{fig:frontier_ppl_2p7b_gpu_mixed}.
  $2{:}4$ baselines are excluded because the CPU does not support $2{:}4$ instructions.}
  \label{fig:frontier_ppl_2p7b_cpu_mixed}
\end{figure*}
We first compare the quality--resource trade-offs achieved under parameter-density, latency, and memory constraints.
Figure~\ref{fig:frontier_ppl_2p7b_gpu_mixed} presents GPU Pareto frontiers for perplexity and mean zero-shot accuracy across all three resources.
Figure~\ref{fig:frontier_ppl_2p7b_cpu_mixed} presents the corresponding CPU frontiers.
The quality metrics give broadly consistent method orderings. 
Under a density budget, unstructured magnitude and Wanda preserve the most quality per retained parameter, but are slower and use much more memory than the dense model.

On the GPU, HSP improves the frontier associated with its optimisation objective. HSP-Lat achieves frontier quality--latency trade-offs, whereas HSP-Mem achieves the best quality--memory trade-offs. 
These gains do not necessarily transfer across objectives; 
HSP-Mem candidates that achieve the best quality--memory trade-offs are slower than the dense model on GPU despite HSP-Lat candidates achieving both frontier quality--latency trade-offs and strong memory reductions.
Even for density constrained methods such as Wanda and Magnitude with $2{:}4$ structures, memory reductions did not translate to equivalent latency reductions.
FLAP and WandaSP reduce latency and memory, but have worse model quality than each relevant HSP method at comparable cost.

On the CPU, HSP-Lat again improves the quality--latency frontier while also achieving the best quality--memory trade-offs.
In contrast, HSP-Mem does not clearly improve the quality--memory frontier on the CPU.
The absence of memory-efficient $2{:}4$ options from the CPU search may contribute to this difference from the GPU results.

Together, these results show that quality--resource trade-offs depend on both the chosen objective and the sparse operations supported by the target device.
No single structure or density-based proxy is uniformly effective, motivating the need for a heterogeneous search space and hardware-aware constraints.
The choice of hardware-aware constraint then should take into account the bottleneck of the target deployment device.

\subsection{Structure Allocation Across Sublayers and Projections}
\label{sec:structures}

\begin{figure*}[!t]
  \centering
  \includegraphics[width=\textwidth]{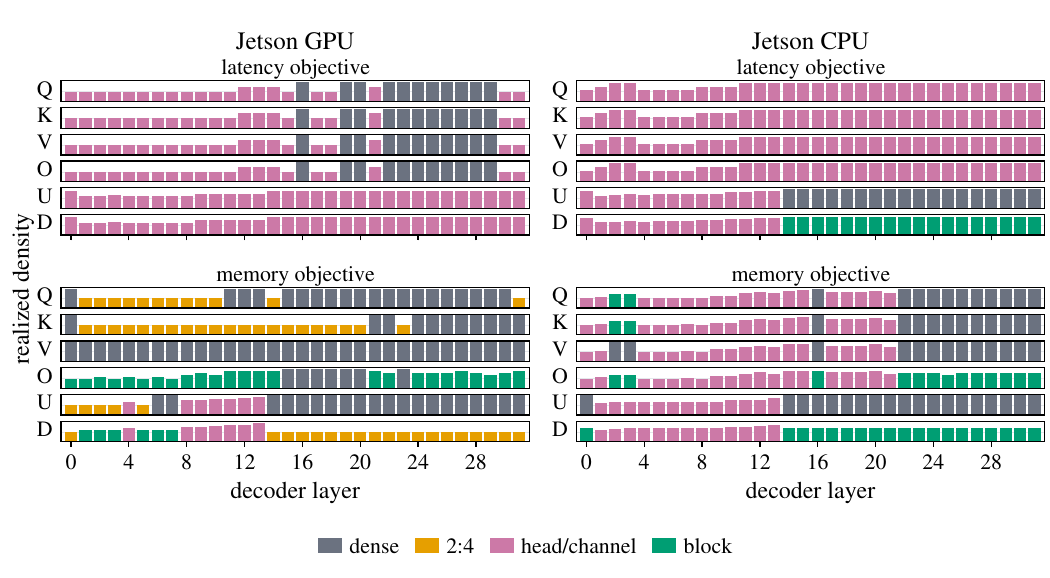}
  \caption{
    Structures selected by HSP for OPT-2.7B on the Jetson Orin Nano Super
    under the mixed workload at a budget ratio of $0.8$.
    Columns are the two devices and rows are the two objectives, so the panels show the
    GPU latency, CPU latency, GPU memory, and CPU memory objectives.
    Within each panel, rows correspond to the attention and MLP projections, while bars correspond to decoder layers.
    Bar height encodes retained density, and colour encodes the selected structure.
  }
  \label{fig:structures_0p8_budget}
\end{figure*}

Figure~\ref{fig:structures_0p8_budget} shows the projection structures and densities that HSP selects for OPT-2.7B at a $0.8$ budget ratio under the latency and memory objectives on the GPU and CPU.
All four candidates share the same target budget hyperparameter, but the selected sublayer options are substantially different.
Two patterns hold in all settings: each candidate combines multiple sparsity structures across its projections, and retained densities vary by decoder layer and projection role, even though the solver can select options that assign one uniform structure and density throughout the model.
On the GPU, the latency objective favours coupled H/C pruning because narrowing the associated projections yields greater speedups with less quality loss at moderate sparsity.
By contrast, the memory objective selects sublayer options that keep $90$ of $192$ projections dense and apply $2{:}4$ to most of the remainder.
On the CPU, where there is no hardware support for $2{:}4$ sparsity, the selected sublayer options instead combine coupled H/C pruning with block sparsity under both objectives and keep deeper MLP up-projections dense.
These allocations demonstrate that the deployment objective and device, rather than the budget level alone, govern structure selection and validate the need for a heterogeneous search space.

\subsection{On-Device Evaluation}

\begin{table*}[!t]
  \centering
  \small
  \setlength{\tabcolsep}{5pt}
  \begin{tabular}{lc@{\hspace{14pt}}cccccc}
    \toprule
    \multirow{2}{*}{Method} & \multirow{2}{*}{PPL}
    & \multicolumn{2}{c}{Prefill}
    & \multicolumn{2}{c}{Decode}
    & \multicolumn{2}{c}{Mixed} \\
    \cmidrule(lr){3-4}\cmidrule(lr){5-6}\cmidrule(lr){7-8}
    & & Mem & Lat & Mem & Lat & Mem & Lat \\
    \midrule
    \multicolumn{8}{c}{\textbf{GPU}} \\
    \midrule
    Dense         & 14.32*           & \multicolumn{6}{c}{OOM}                                                                             \\
    Wanda-$2{:}4$ & 21.90            & \textbf{3447}   & 0.91            & \textbf{3307}   & 8.80            & \textbf{3748}   & \underbar{9.62} \\
    FLAP          & 17.15            & 4519            & \underbar{0.64} & 4158            & \underbar{6.69} & 4519            & 9.97            \\
    HSP-Lat       & \underbar{15.31} & \underbar{4367} & \textbf{0.62}   & \underbar{4065} & \textbf{6.52}   & \underbar{4367} & \textbf{9.12}   \\
    HSP-Mem       & \textbf{15.15}   & 4519            & 11.56           & 4182            & 18.48           & 4621            & 32.86           \\
    \midrule
    \multicolumn{8}{c}{\textbf{CPU}} \\
    \midrule
    Dense         & 15.66            & 6201            & 84.78            & 5676            & 27.24            & 6135            & 117.76           \\
    Wanda-$2{:}4$ & \multicolumn{7}{c}{No supported $2{:}4$ kernel}                                                                                            \\
    FLAP          & 18.23            & \textbf{6512}   & \textbf{66.55}   & \textbf{6088}   & \textbf{22.82}   & \textbf{6570}   & \textbf{95.54}   \\
    HSP-Lat       & \textbf{15.50}   & 6768            & \underbar{67.03} & \underbar{6425} & 24.64            & 6748            & \underbar{97.16} \\
    HSP-Mem       & \underbar{17.68} & \underbar{6648} & 71.54            & 6550            & \underbar{22.97} & \underbar{6640} & 100.02           \\
    \bottomrule
  \end{tabular}
  \caption{On-device evaluation of OPT-2.7B on the Jetson Orin Nano Super GPU and CPU.
    Peak memory ($\downarrow$) is in megabytes (MB) and end-to-end latency ($\downarrow$) in seconds (s).
    Best and second-best pruned candidates within each device are \textbf{bold} and \underbar{underlined} respectively.
    Pruned candidates use a budget equal to $80\%$ of the dense cost under their respective constraints, except Wanda-$2{:}4$ which only supports a density constraint of $50\%$.
    The dense model exceeds GPU memory due to runtime overhead, so its perplexity in the GPU block is measured off-device ($*$) and its GPU costs are reported as OOM.
    The CPU runtime has no supported $2{:}4$ kernel.
    CPU perplexity uses only the first 32 WikiText-2 windows of 1024 tokens each and is not directly comparable to the perplexities in the GPU block.}
  \label{table:edge_eval}
\end{table*}

We evaluate models on the target device to compare measured resource use with the modelled resource estimates.
We run OPT-2.7B on an NVIDIA Jetson Orin Nano Super at a fixed $25\,\mathrm{W}$ power mode, reporting on-device peak memory and end-to-end latency for different workloads (Table~\ref{table:edge_eval}).
Unstructured candidates are excluded as they exceed memory capacity.

These results show that hardware efficiency is objective-specific. 
On the GPU, HSP-Lat is consistently the fastest candidate, whereas the structures selected under the memory constraint reduce storage but lead to higher latency. 
On the CPU, HSP-Lat provides a stronger quality--latency trade-off than FLAP and improves prefill latency over the dense model. 
The agreement between the predicted and measured trade-offs supports using the offline models to select pruning configurations for the deployment bottleneck of interest.

\subsection{Ablation of Search Space and Resource Proxy}

\begin{table}[h]
  \centering
  \setlength{\tabcolsep}{1.2pt}
  {\small
    \begin{tabular}{llcccccc}
      \toprule
      & & \multicolumn{2}{c}{Prefill} & \multicolumn{2}{c}{Decode} & \multicolumn{2}{c}{Mixed} \\
      \cmidrule(lr){3-4}\cmidrule(lr){5-6}\cmidrule(lr){7-8}
      Search & Constraint & PPL  & Lat  & PPL  & Lat  & PPL  & Lat  \\
      \midrule
      Dense &  & 14.32 & 0.755 & 14.32 & 9.890 & 14.32 & 11.7 \\
      \midrule
      Single & Density & 18.40 & \textbf{0.525} & \underline{18.40} & \underline{7.530} & \underline{18.40} & 8.995 \\
      Hetero. & Density & \textbf{13.73} & 85.5 & \textbf{13.73} & 31.0 & \textbf{13.73} & 117.5 \\
      Single & Latency & 18.85 & \underline{0.537} & 20.95 & \textbf{7.045} & 22.61 & \underline{8.292} \\
      Hetero. & Latency & \underline{16.62} & \underline{0.537} & 20.75 & \textbf{7.045} & 21.28 & \textbf{8.291} \\
      \bottomrule
    \end{tabular}
  }
  \caption{Ablation of search space heterogeneity and constraint type for OPT-2.7B on the Jetson Orin Nano Super GPU with $B=0.7$. \textit{Single} uses coupled H/C options, while \textit{Hetero.}\ uses the full heterogeneous search space. PPL ($\downarrow$) is WikiText-2 perplexity and Lat ($\downarrow$) is profile-backed latency in seconds (s). Marking follows Table~\ref{table:edge_eval}.}
  \label{table:ablation_ppl_latency_0p7}
\end{table}

In Table~\ref{table:ablation_ppl_latency_0p7}, we separately ablate the heterogeneous search space and the hardware-aware constraint to show that both are necessary.
A heterogeneous search space with a density constraint achieves the best perplexity (13.73) but has two orders of magnitude higher profile-backed latency than that of the dense model, reinforcing the notion that density is not a suitable proxy for latency.
H/C pruning with a latency constraint (\textit{Single--Latency}) meets the budget but degrades quality because it restricts the solver to a narrower set of latency-efficient configurations.
Our complete method (\textit{Hetero.--Latency}) moves the quality--latency frontier by meeting the latency budget while retaining higher model quality.
Neither component suffices alone: heterogeneity without a hardware-aware constraint yields the best quality at prohibitive deployment cost, but a latency constraint over a single structure meets the budget by sacrificing quality.

%% file: sections/conclusion.tex
\section{Conclusion}
In this work, we propose Heterogeneous Structured Pruning (HSP), a training-free pruning framework in which a global solver selects one option for every attention and MLP sublayer under a deployment budget, thereby assigning a sparsity structure and density to each projection.
Instead of optimising a hardware-agnostic density ratio, HSP assigns every candidate structure and density an explicit deployment cost: latency obtained from reusable kernel profiles on the target device or memory computed analytically from the storage formats and the workload-dependent key--value cache.
The model-wide allocation is then solved as a multiple-choice knapsack problem (MCKP).
Density floors exclude extreme candidates for which the importance proxy becomes unreliable.
On OPT-2.7B and an NVIDIA Jetson Orin Nano Super, HSP-Lat improves the quality--latency frontier on both the CPU and GPU, while HSP-Mem improves the GPU quality--memory frontier.
These frontiers use modelled resource costs, with end-to-end latency and peak memory evaluated separately on device.
The selected configurations show that both the target hardware and the resource objective govern which sparsity structures are effective.
By combining a heterogeneous search space with hardware-aware resource constraints, HSP enables pruning to be tailored to the target device and its deployment bottleneck.

%% file: sections/acknowledgements.tex
\section*{Acknowledgements}

We acknowledge support of the UKRI AI programme, and the Engineering and Physical Sciences Research Council, for APRIL---EPSRC AI Hub for Productive Research \& Innovation in electronics [grant number EP/Y029763/1].
Cameron Barker is supported by the UKRI EPSRC Centre for Doctoral Training in Machine Learning Systems [EP/Y03516X/1].
This publication is part of the LESSEN project (NWA.1389.20.183), funded in part by the Dutch Research Council (NWO) through the NWA ORC 2020/21 programme.
Henry Gouk is supported by the Royal Academy of Engineering under its Research Fellowship programme.

%% file: sections/appendix/experimental-configuration.tex
\section{Experimental Configuration}
\label{app:experimental_configuration}

\subsection{Option and Hyperparameter Configuration}

We use the independent and coupled option families defined in
Section~\ref{sec:option_sets}, with density settings summarised in
Table~\ref{table:app_candidate_settings}.

\begin{table}[H]
  \centering
  \small
  \setlength{\tabcolsep}{4pt}
  \renewcommand{\arraystretch}{1.15}
  \begin{tabular}{@{}p{0.20\linewidth}p{0.27\linewidth}p{0.16\linewidth}p{0.28\linewidth}@{}}
    \toprule
    Structure & Requested densities & OPT floor & Exceptions \\
    \midrule
    Dense & $1$ & -- & Always available \\
    Unstructured & $\{0.1,0.2,\ldots,0.9\}$ & $0.3$ & -- \\
    Rectangular blocks & $\{0.1,0.2,\ldots,0.9\}$ & $0.5$ & CPU memory: floor $0.7$ \\
    H/C & Profiled widths, increments of $1/H$ & $0.5$ & LLaMA 7B: floor $0.125$ \\
    $2{:}4$ & $0.5$ (fixed) & -- & GPU only \\
    \bottomrule
  \end{tabular}
  \caption{Candidate density settings. Requested densities below the applicable
    floor are excluded. $H$ is the number of attention heads. Coupled options
    enumerate every retained head count and every MLP width on the profiling
    grid at or above the floor.}
  \label{table:app_candidate_settings}
\end{table}

Block shapes are output rows $\times$ input columns. The GPU grid is
$\{1,16,32,64\}\times\{1,16,32,64\}$, and the CPU grid is
$\{1,8,16,32\}\times\{1,8,16,32\}$.
The $1\times1$ case is represented by the unstructured option, and every other
pair is block sparse. We exclude $2{:}4$ from the CPU search because it has no
specialised hardware support.

For an unstructured option, each output row retains the nearest integer to the
requested fraction of its weights.
For a block-sparse option, each block-grid row similarly retains the nearest
integer number of blocks.
Half-integer ties are rounded upward in both cases.

\paragraph{Hyperparameter Development}
The density-floor motivation is described in Section~\ref{sec:option_sets}.
We swept minimum densities on a development model and workload using WikiText-2 validation perplexity, then chose values that worked across settings rather than tuning every experiment separately.

\paragraph{Coupled-Option Loss Calibration}
We permit an option-specific weight on the outer loss of each inner-problem solution.
In an early development experiment, we compared $w\in\{1.0,0.95,0.85,0.7,0.5\}$ for the coupled-option solver loss.

The value $w=0.95$ gave a small validation-perplexity improvement for
latency-constrained searches. Accordingly, coupled options use weight $0.95$
in the latency experiments and $1$ in the memory experiments. Independent
options have weight $1$ in both cases.

\subsection{Hardware and Evaluation Setup}

Table~\ref{table:app_experimental_setup} summarises the hardware, software,
and calibration settings.

\begin{table}[H]
  \centering
  \small
  \setlength{\tabcolsep}{4pt}
  \renewcommand{\arraystretch}{1.1}
  \begin{tabular}{@{}p{0.25\linewidth}p{0.71\linewidth}@{}}
    \toprule
    Setting & Configuration \\
    \midrule
    Target device & NVIDIA Jetson Orin Nano Super \\
    CPU & Six-core ARM Cortex-A78AE with NEON \\
    GPU & 1024 CUDA cores and 32 Tensor Cores \\
    Shared memory & 8 GB LPDDR5, peak bandwidth 102.4 GB/s \\
    Software & NVIDIA JetPack 6.2.1, L4T 36.4.7, CUDA 12.6, PyTorch 2.5.0 (prebuilt package) \\
    Device controls & 25 W mode, fans at 100\%, $55\,^{\circ}\mathrm{C}$ temperature gating between profiling measurements \\
    \addlinespace
    Pruning server & Four NVIDIA L40 GPUs, each with 48 GB memory \\
    Calibration & Shared by all activation-dependent methods: 256 randomly selected, contiguous 1024-token windows from WikiText-2 training data \\
    Evaluation harness & EleutherAI's \texttt{lm-evaluation-harness} v0.4.12\newline
      \citep{gao2024lm_eval}, Hugging Face model adapter \\
    Zero-shot seeds & Python, NumPy, PyTorch and few-shot harness seeds all set to 0 \\
    \bottomrule
  \end{tabular}
  \caption{Experimental setup. Pruning and search run offline on the server.
    On-device measurements use the Jetson Orin Nano Super.}
  \label{table:app_experimental_setup}
\end{table}

During pruning and search, dense operations and explicit masks simulate pruning without using sparse kernels.
Reported latency and memory values are obtained from the corresponding resource models unless they are specified as ``on-device'' measurements,
which are measured directly on the NVIDIA Jetson Orin Nano Super.

\subsection{Budgets, Metrics, and Randomness}

\paragraph{Resource Budgets}
Latency- and memory-constrained searches use a fixed budget $B$ in units of time or bytes, respectively.
The solver allocates $B-C_0$ across sublayers. Reported total costs include $C_0$.
For memory, the sublayer costs include projection weights and biases, sparse metadata, and peak key--value cache size.
For latency, they include the profiled projection and attention operations.
Density-constrained baselines instead target prunable projection weights, following the convention in Section~\ref{sec:structured_pruning}.
They do not impose a latency or encoded-memory budget.

\paragraph{Evaluation Metrics}
We report perplexity on the WikiText-2 test split~\citep{merity2017pointer}.
Its text rows, including empty rows, are joined with two newlines and tokenised
once with the evaluated model's tokenizer.
We then use 1024-token windows with stride 1024.
Due to runtime constraints, the on-device CPU perplexity results in the main
paper use only the first 32 windows.
All other perplexity results use the full test split.

We also report the unweighted mean zero-shot accuracy across BoolQ, PIQA,
HellaSwag, WinoGrande, ARC-e, ARC-c, and
OpenBookQA~\citep{clark2019boolq,Bisk2020,zellers2019hellaswag,sakaguchi2019winogrande,allenaiarc,OpenBookQA2018}.
Following the harness, we use plain accuracy for BoolQ and
WinoGrande and length-normalised accuracy for the other five tasks.

The resource metrics are retained parameter density, profile-backed latency,
and analytical memory including the workload-dependent key--value cache.
In figures, we normalise each to the corresponding dense model unless absolute units are
stated.

\paragraph{Randomness}
Given the calibration statistics in Table~\ref{table:app_experimental_setup}, mask construction, dominance filtering,
and MCKP selection are deterministic.
Perplexity evaluation is deterministic.
Kernel timing and cache control follow the measurement protocol in
Section~\ref{app:measurement_protocol}.

%% file: sections/appendix/procedure.tex
\FloatBarrier
\section{HSP Implementation and Solver}
\label{app:solver_and_realisation}

\subsection{Candidate Construction and Scoring}
\label{app:option_construction}

\paragraph{Independent Options}
Let $\Phi^{(\pi)}$ be the finite set of structure-density choices for projection $\pi$ after applying the density floors and device restrictions in Section~\ref{app:experimental_configuration}.
An option $\phi^{(\pi)}=(s^{(\pi)},d^{(\pi)})$ selects one such pair.
The independent sublayer option set and its admissible mask sets are Cartesian
products over projections, combining one option or mask from each projection:
\begin{equation}
  \begin{aligned}
    \Phi_{\mathrm{ind}}^{(l)}&=\prod_{\pi\in\Pi_l}\Phi^{(\pi)},\\
    \mathcal{M}^{(l)}(\phi^{(l)})
    &=\prod_{\pi\in\Pi_l}\mathcal{M}^{(\pi)}(s^{(\pi)},d^{(\pi)}).
  \end{aligned}
  \label{eq:independent_options}
\end{equation}
Each projection's mask is selected independently using
Equation~\ref{eq:projection_selection}.

\paragraph{Coupled Options}
Let $\mathcal{G}_{\mathrm{cpl}}^{(l)}$ contain the admissible shared densities for a coupled sublayer, and let $s_{\mathrm{cpl}}^{(\pi)}$ be the row- or column-slice structure required by projection $\pi$'s role.
Then
\begin{equation}
  \Phi_{\mathrm{cpl}}^{(l)}
  =\left\{\bigl(s_{\mathrm{cpl}}^{(\pi)},d\bigr)_{\pi\in\Pi_l}
    :d\in\mathcal{G}_{\mathrm{cpl}}^{(l)}\right\}.
  \label{eq:coupled_options}
\end{equation}
For the standard multi-head attention and two-projection MLP in Section~\ref{sec:structured_pruning}, admissible masks satisfy
$\mathbf{M}^{(\pi)}=(\mathbf{M}^{(\pi_{\mathrm{c}})})^{\top}$ for each producer $\pi\ne\pi_{\mathrm{c}}$.
The consumer mask belongs to
$\mathcal{M}^{(\pi_{\mathrm{c}})}(s_{\mathrm{cpl}}^{(\pi_{\mathrm{c}})},d)$
and retains complete heads or hidden channels.
The inner problem selects this mask using only the consumer's importance scores.
The producer masks follow from the coupling constraints and are included in the outer loss.
Gated MLP and grouped-query attention extensions are described in Section~\ref{app:coupled_structures}.

Together, these families form the disjoint union
$\Phi^{(l)}=\Phi_{\mathrm{ind}}^{(l)}\uplus\Phi_{\mathrm{cpl}}^{(l)}$.

\paragraph{Outer-Score Normalisation}
To compare candidates across projections, we normalise each projection's
outer importance scores by their mean. For a positive mean,
\begin{equation}
  \widehat{\mathbf{S}}^{(\pi)}
  =\frac{\mathbf{S}^{(\pi)}}{\overline{S}^{(\pi)}},
  \qquad
  \overline{S}^{(\pi)}
  =\frac{\sum_{i,j}S^{(\pi)}_{i,j}}{\operatorname{numel}(\mathbf{W}^{(\pi)})}.
  \label{eq:importance_normalisation}
\end{equation}
An all-zero score matrix is left unchanged.
The normalised scores enter the outer loss in Equation~\ref{eq:outer_loss},
with total importance equal to the number of weights for each nonzero projection.
Inner mask construction uses the base scores of the chosen pruning method.
Scaling these scores by a positive projection-wise constant leaves the inner
mask selections unchanged.

\subsection{Independent Option Mask Construction}
\label{app:independent_masks}

For each independent option, the sparsity structure defines the comparison
groups in the inner problem of Equation~\ref{eq:projection_selection}.
Weights or blocks are ranked by importance within each group, and the
requested density determines how many are retained.

Unstructured masks keep the highest-scoring weights in each output row.
For $2{:}4$, we keep the two highest-scoring weights in each group of four
consecutive input columns within a row.
For rectangular blocks, we sum the importance scores within each block, then
keep the highest-scoring blocks in each row of the block grid.

For unstructured and block masks, the requested density determines how many
weights or blocks to retain per row, using the rounding rule in
Section~\ref{app:experimental_configuration}.
With these counts fixed, selecting the highest-scoring weights or blocks
minimises the total importance removed.

\subsection{MCKP Solver}
\label{app:mckp_solver}

For each option $\phi\in\Phi^{(l)}$, write the outer loss of its constructed mask as
$e_l(\phi)=\ell_{\mathrm{out}}^{(l)}(\mathbf{M}^{\star(l)}(\phi);\phi)$.
The solver selects one loss--cost pair $(e_l(\phi),C_l(\phi))$ per sublayer
under the budget $B-C_0$, as in Equation~\ref{eq:hsp_objective}.

Before allocation, we apply the projection- and sublayer-level dominance
filtering described in Section~\ref{sec:hsp_bilevel}.

\paragraph{Dynamic Programming}
We solve the allocation by dynamic programming~\citep{kellerer2004knapsack},
building configurations one sublayer at a time and retaining a frontier
$\mathcal{F}$ of cumulative loss--cost pairs.
To limit its size, we group costs into bins of width $\Delta$ and keep the
lowest-loss state in each bin before removing dominated states.
We use $\Delta=0.01\,\mathrm{ms}$ for latency and approximately $1.05\,\mathrm{MB}$ for memory.
Budget checks use unrounded costs, and the final configuration is checked
against $B$, including the fixed cost $C_0$.
Unlike dominance filtering, bin merging can discard an optimum, so the solver
is approximate.

\subsection{HSP Algorithm}

\begin{algorithm}[H]
  \caption{Heterogeneous Structured Pruning}
  \label{alg:hsp}
  \begin{algorithmic}[1]
    \REQUIRE Weights $\mathbf{W}$; calibration data $\mathcal{D}_{\mathrm{cal}}$;
    option sets $\{\Phi^{(l)}\}_{l=1}^{L}$; resource costs $C$; absolute budget $B$
    \ENSURE Selected configuration $\boldsymbol{\phi}^{\star}$ and pruned weights
    \STATE Compute inner importance scores and mean-normalised outer scores
    \FOR{$l=1$ \TO\ $L$}
      \STATE Construct each candidate mask $\mathbf{M}^{\star(l)}(\phi)$
      using the inner problem
      \STATE Evaluate its outer loss $e_l(\phi)$ and resource cost $C_l(\phi)$
      \STATE Pareto-filter the options using $(e_l(\phi),C_l(\phi))$
    \ENDFOR
    \If{$C_0+\sum_l\min_{\phi\in\Phi^{(l)}}C_l(\phi)>B$}
      \STATE \textbf{return} infeasible
    \EndIf
    \STATE $\mathcal{F}\leftarrow\{(\varnothing,0,0)\}$
    \FOR{$l=1$ \TO\ $L$}
      \STATE Extend $\mathcal{F}$ by every option in $\Phi^{(l)}$
      \STATE Discard states whose unrounded cost exceeds $B-C_0$
      \STATE Keep one representative per $\Delta$-bin, then Pareto-filter
    \ENDFOR
    \STATE Reconstruct $\boldsymbol{\phi}^{\star}$ from the lowest-loss final state
    \STATE Apply the selected masks and configured bias compensation
    \STATE \Return Pruned weights and $\boldsymbol{\phi}^{\star}$
  \end{algorithmic}
\end{algorithm}

\subsection{Coupled-Structure Extensions}
\label{app:coupled_structures}

The coupled masks extend to gated MLPs and grouped-query attention (GQA).

\paragraph{Gated MLPs}
A gated MLP adds a gate projection $G$ alongside the up projection $U$.
We apply the same retained-channel row mask to $U$ and $G$ and the
corresponding column mask to the down projection $D$, so
$\mathbf{M}^{(G)}=\mathbf{M}^{(U)}=(\mathbf{M}^{(D)})^{\top}$.
The inner loss remains the consumer-projection loss on $D$, while the option
cost includes $U$, $G$, and $D$.

\paragraph{Grouped-Query Attention}
In GQA, several query heads share one key--value head.
Removing a query head removes its $d_h$ rows of $Q$ and the matching $d_h$
columns of $O$.
We keep a key--value head, and its $d_h$ rows of $K$ and $V$, if and only if
we keep at least one of the query heads that share it.
The $Q$ mask is therefore still the transpose of the $O$ mask, but the $K$ and
$V$ masks are not.
A GQA option keeps $d_QH_Q$ of the $H_Q$ query heads, chosen so that they
share exactly $d_{\mathrm{KV}}H_{\mathrm{KV}}$ of the $H_{\mathrm{KV}}$
key--value heads.
Fixing both head counts fixes the option's cost, including the key--value
cache, before the inner problem chooses which heads to keep.

\subsection{Bias Compensation}
Consider a projection
$\mathbf{y}=\mathbf{W}\mathbf{x}+\mathbf{b}$ with mask $\mathbf{M}$.
Let $\bar{\mathbf{x}}$ be the mean of the rows of the input-activation matrix
$\mathbf{X}^{(\pi)}$ over the calibration tokens. We use the compensated bias
\begin{equation}
  \widetilde{\mathbf{b}}
  =
  \mathbf{b}
  +
  \left(\mathbf{W}-\mathbf{W}\odot\mathbf{M}\right)\bar{\mathbf{x}},
  \label{eq:app_bias_compensation}
\end{equation}
which matches the original projection output at the calibration mean.
For HSP, we apply Equation~\ref{eq:app_bias_compensation} to the consumer
projection ($O$ or $D$) of each coupled option and to every independently
block-sparse projection, using that projection's input mean.
We do not compensate coupled producer projections or independent unstructured
and $2{:}4$ masks.
All reported OPT HSP runs enable this correction and modify their existing
biases, so no additional storage or latency is introduced.
The bias-free LLaMA-7B HSP runs do not use bias compensation.

%% file: sections/appendix/sparsity-support.tex
\section{Resource Models, Kernels and Measurement}
\label{app:sparsity_hardware}

\subsection{Storage Formats and Kernels}
\label{app:kernels}

Sparse compilers such as TACO generate kernels for tensor algebra
\citep{kjolstad2017taco}, but the implementations we tested did not provide
sufficient performance across our operator shapes and moderate densities.
We therefore supplement available operators with our own CPU and CUDA kernels
for the Jetson Orin Nano Super.
Table~\ref{table:app_storage_kernels} summarises the storage formats and kernel
families used for each structure.

\begin{table}[H]
  \centering
  \small
  \setlength{\tabcolsep}{4pt}
  \renewcommand{\arraystretch}{1.15}
  \begin{tabular}{@{}p{0.21\linewidth}p{0.23\linewidth}p{0.21\linewidth}p{0.27\linewidth}@{}}
    \toprule
    Structure & Weight storage & CPU kernels & GPU kernels \\
    \midrule
    Dense & Dense matrix & Custom C++ & PyTorch dense \\
    H/C & Narrowed dense matrix & Custom C++ & PyTorch dense \\
    Unstructured & CSR & Custom C++  & Custom CUDA \\
    Rectangular blocks & BSR & Custom C++ & Custom CUDA \\
    $2{:}4$ & NV24 & Not supported & \raggedright CUTLASS and\\Custom CUDA \tabularnewline
    \bottomrule
  \end{tabular}
  \caption{Projection storage and kernel families on the Jetson Orin Nano Super.
    H/C pruning uses dense kernels at reduced dimensions. Compatible kernels
    are compared during profiling.}
  \label{table:app_storage_kernels}
\end{table}

Compressed sparse row (CSR) and block sparse row (BSR) store retained values
or blocks with column indices and row pointers. NV24 stores half the original
values with bit-packed position metadata. These metadata costs make storage
depend on the structure as well as the retained density
(Figure~\ref{fig:structure_memory}).

All deployed kernels use FP16 weights and activations with FP32 accumulation.
The GPU pool also includes PyTorch attention operators. Sparse Tensor Cores
accelerate $2{:}4$, and the CUDA block kernels use Tensor Cores when their
structure and dimensions permit. The CPU kernels use multithreaded C++ with
ARM NEON. PyTorch's dense FP16 CPU kernels were included in the sweep, but our
implementations were substantially faster and form the final CPU profiles.
We release our CPU and CUDA kernels to support further research.

\subsection{Resource Models}

\paragraph{Analytic Memory Model}

We count FP16 weights and biases, sparse metadata, and the final key--value
cache. For an $m\times n$ weight matrix, storage costs in bytes are
\begin{equation}
  \begin{aligned}
    C_{\mathrm{dense}} &= 2mn,\\
    C_{\mathrm{BSR}} &= 2zrc + 4z + 4\left(\frac{m}{r}+1\right),\\
    C_{2{:}4} &= \frac{9mn}{8},
  \end{aligned}
  \label{eq:app_projection_memory}
\end{equation}
where BSR stores $z$ retained $r\times c$ blocks with 32-bit column indices
and row pointers. CSR is the case $r=c=1$.
The $2{:}4$ format stores half the weights and two metadata bytes per 16
original weights. H/C pruning uses the dense formula at the reduced dimensions.

For batch size one, the FP16 key--value cache costs
\begin{equation}
  C_{\mathrm{KV}}
  =4T\sum_{k}H_{\mathrm{KV}}^{(k)}d_h^{(k)},
  \label{eq:app_kv_memory}
\end{equation}
where $T$ is the final cache length (prefill plus decode tokens), and
$H_{\mathrm{KV}}^{(k)}$ and $d_h^{(k)}$ are the retained key--value head
count and head dimension in decoder layer $k$. The factor 4 accounts for two bytes
each for keys and values.

Total memory adds projection storage, retained biases, the cache, and fixed
parameter storage $C_0$, such as embeddings and layer norms. Projection biases belong
to the projection costs, and shared parameters such as tied embedding and
output-head weights are counted once.
This model excludes temporary activations, allocator overhead, and kernel
workspaces. Measured peak memory is reported separately in the on-device results.

\paragraph{Profile-Backed Latency Model}
For latency-constrained pruning, we assign each candidate the summed measured
latencies of its induced projection and attention operations rather than using
parameter count or FLOPs as a runtime proxy. The kernel pool is listed in Table~\ref{table:app_storage_kernels}.
Sections~\ref{app:kernel_profiles} and~\ref{app:measurement_protocol} describe
profile lookup and measurement.

\subsection{Kernel Profiles}
\label{app:kernel_profiles}

We profile every projection and attention operation induced by the HSP-Lat
option space on the NVIDIA Jetson Orin Nano Super in its fixed $25\,\mathrm{W}$
power mode. For each operation, the sweep measures compatible implementations
and a curated set of launch configurations using one fixed set of random FP16
inputs, weights, and structure-compatible masks per profiling job, then retains
the implementation with the lowest median latency.

Each induced operation is matched exactly by device, inference phase,
matrix dimensions, sparsity structure, and retained density where applicable.
We profile full coverage of the search space and do not interpolate or extrapolate missing entries.
For a mixed workload, prefill is counted once and decode once per
generated token.

\begin{figure*}[!t]
  \centering
  \includegraphics[width=\textwidth]{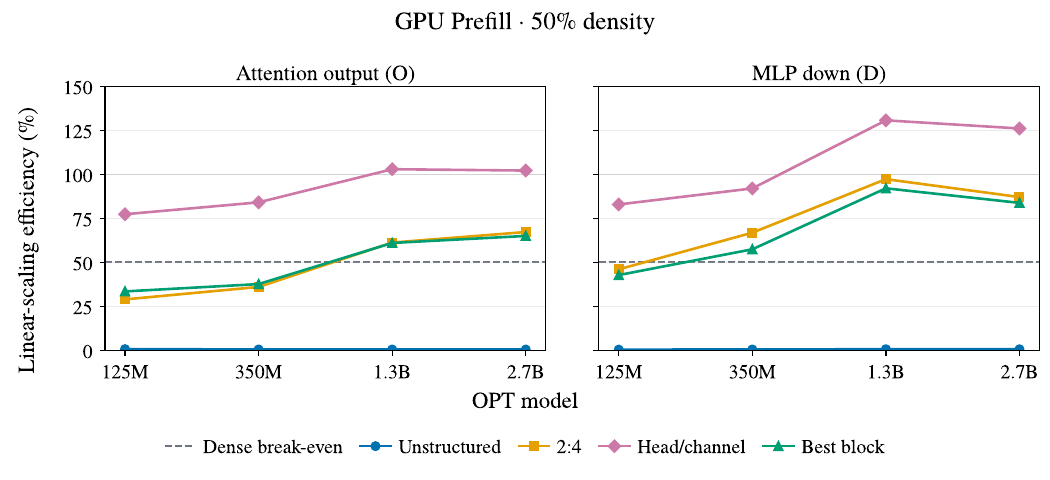}
  \caption{GPU-prefill linear-scaling efficiency at retained density $d=0.5$
    across the OPT model family for the attention output (O) and MLP down (D)
    projections. ``Best block'' is the fastest profiled rectangular block
    shape for each model, which is $64\times64$ in every case shown. The dashed
  $50\%$ line marks break-even with dense execution.}
  \label{fig:kernel_efficiency_model_scale}
\end{figure*}

\begin{figure*}[t]
  \centering
  \begin{minipage}[t]{0.49\textwidth}
    \vspace{0pt}
    \centering
    \includegraphics[width=\linewidth]{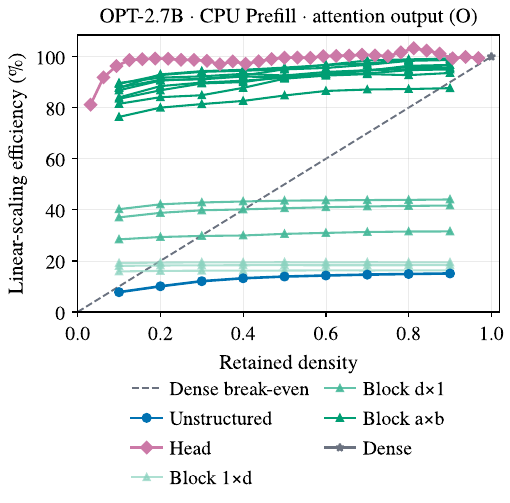}
    \caption{Linear-scaling efficiency across retained densities for the
      OPT-2.7B attention output (O) projection during CPU prefill. The CPU
      profiles exclude $2{:}4$, for which the CPU has no dedicated hardware
      primitive. Points above the dashed $100d\%$ line are faster than dense
    execution.}
    \label{fig:kernel_efficiency_cpu_o}
  \end{minipage}
  \hfill
  \begin{minipage}[t]{0.49\textwidth}
    \vspace{0pt}
    \centering
    \includegraphics[width=\linewidth]{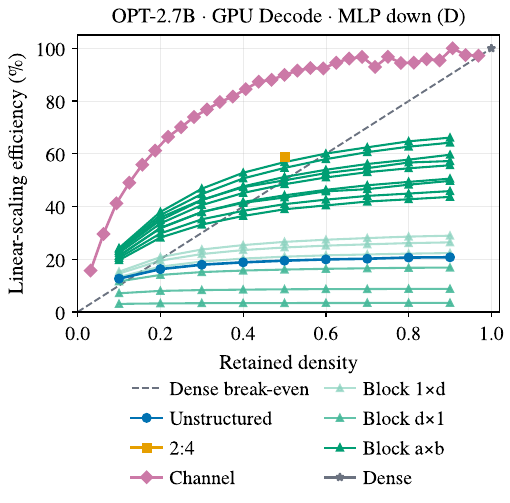}
    \caption{Linear-scaling efficiency across retained densities for the
      OPT-2.7B MLP down (D) projection during one-token GPU decode. Points
    above the dashed $100d\%$ line are faster than dense execution.}
    \label{fig:kernel_efficiency_gpu_decode_d}
  \end{minipage}
\end{figure*}

Figure~\ref{fig:kernel_efficiency_qoud} shows GPU-prefill profiles for OPT-2.7B.
Figures~\ref{fig:kernel_efficiency_model_scale}--\ref{fig:kernel_efficiency_gpu_decode_d}
extend these measurements across model sizes, devices, and inference phases.
All use the linear-scaling efficiency defined in Figure~\ref{fig:kernel_efficiency_qoud}.

Sparse GPU kernels generally become more efficient as projection dimensions
grow, whether through wider MLP projections or larger models. H/C pruning
remains strong across sizes, while $2{:}4$ and block sparsity become more
competitive on larger projections. A structure may therefore speed up an MLP
projection but not an attention projection because of kernel performance at
different matrix sizes. This means structures can be chosen based on their
performance characteristics not just by differences in weight importance structure.
Their gains also depend on density and
inference phase. On the CPU, H/C and several block shapes approach ideal
linear scaling, while unstructured sparsity remains slower than dense execution
in the profiles shown.

Statically optimising kernels to a fixed post-pruning pattern could improve efficiency,
particularly for unstructured and block sparsity.
We do not consider such model-specific optimisation,
so the profiled kernels could be improved in the future.

\subsection{Measurement Protocol}
\label{app:measurement_protocol}

All profiles and on-device evaluations use the fixed power, fan, and thermal
controls in Section~\ref{app:experimental_configuration}. For each kernel configuration, we perform two
warm-ups followed by ten timed repetitions and use the median latency. We wait
for the device temperature to fall below $55\,^{\circ}\mathrm{C}$ and flush the
L2 cache before each iteration. GPU timings use CUDA events and CPU timings use high-precision timers.

For whole-model results, we perform one warm-up followed by five timed GPU
executions or three timed CPU executions and report the median.
We measure peak memory separately from latency. For GPU runs, we report the
maximum memory allocated through the GPU allocator. For CPU runs, we report
the maximum physical RAM used by the process.
Across the twelve HSP-Lat and HSP-Mem device--workload pairs in Table~\ref{table:edge_eval}, the median absolute relative error,
$\lvert L_{\mathrm{meas}}/L_{\mathrm{prof}}-1\rvert$, between summed profile latency and measured end-to-end latency is $2.8\%$.

\clearpage
\subsection{Computational Cost}
\label{app:computational_cost}

\paragraph{Kernel Profile Sweeps}
Profiling OPT-2.7B takes $1.8\,\mathrm{h}$ on the CPU and $0.7\,\mathrm{h}$
on the GPU (Table~\ref{table:profile_sweep_cost}). Profiles can be reused
across budgets, importance metrics, and models when the device, runtime,
operation shapes, inference phases, structures, and densities match.

\begin{table}[!htbp]
  \centering
  \setlength{\tabcolsep}{3pt}
  {\small
    \begin{tabular}{lcccc}
      \toprule
      Model & \multicolumn{2}{c}{Orin Nano CPU} & \multicolumn{2}{c}{Orin Nano GPU} \\
      \cmidrule(lr){2-3}\cmidrule(lr){4-5}
      & Configs & Time (min) & Configs & Time (min) \\
      \midrule
      OPT-125M & 2,004 & 10.4 & 2,964 & 4.7 \\
      OPT-350M & 2,104 & 18.5 & 3,064 & 7.3 \\
      OPT-1.3B & 2,464 & 77.4 & 3,424 & 25.8 \\
      OPT-2.7B & 2,464 & 110.6 & 3,424 & 39.0 \\
      \bottomrule
    \end{tabular}
  }
  \caption{One-time kernel-profiling cost per model on the Jetson Orin Nano. ``Configs'' is the number of measured kernel configurations, and time is their total measurement wall-clock time.}
  \label{table:profile_sweep_cost}
\end{table}

\paragraph{Pruning Wall-Clock Time}
\label{app:pruning_time}

With the model loaded and calibration statistics precomputed, producing an
OPT-2.7B pruning decision takes $8\text{--}22\,\mathrm{s}$ with HSP and
$0.33\text{--}1.15\,\mathrm{s}$ with the baselines
(Table~\ref{table:prune_time}). Constructing candidate masks is the largest
component of HSP's runtime. Kernel development and tuning time is not included
and was not measured.

\begin{table}[!htbp]
  \centering
  \setlength{\tabcolsep}{3pt}
  {\small
    \begin{tabular}{lcccc}
      \toprule
      Method & \multicolumn{4}{c}{Time (s)} \\
      \cmidrule(lr){2-5}
      & Options & Solve & Select & Total \\
      \midrule
      Wanda & -- & -- & -- & 1.15 \\
      FLAP & -- & -- & -- & 0.33 \\
      \midrule
      HSP-Lat (CPU) & 19.22 & 1.93 & 0.89 & 22.05 \\
      HSP-Lat (GPU) & 20.30 & 0.39 & 0.31 & 21.01 \\
      HSP-Mem (CPU) & 4.16 & 0.86 & 3.98 & 9.00 \\
      HSP-Mem (GPU) & 5.44 & 0.79 & 1.75 & 7.99 \\
      \bottomrule
    \end{tabular}
  }
  \caption{Time to produce an OPT-2.7B pruning decision ($\downarrow$), using
    precomputed calibration statistics and excluding model loading. ``Options'' constructs the
    inner-problem solutions, ``Solve'' measures the MCKP runtime, and
  ``Select'' covers mask construction and bias compensation.}
  \label{table:prune_time}
\end{table}

\FloatBarrier

%% file: sections/appendix/additional-results.tex
\Needspace{9\baselineskip}
\section{Additional Results and Ablations}
\label{app:scaling}

We extend the comparisons in Section~\ref{sec:quality_resource_pareto} across inference workloads, OPT model
sizes, and LLaMA 7B, then examine selected block shapes and ablations.
Missing HSP points or panels indicate that no structure was found within the
resource budget.

\subsection{Scaling Across Workloads}

\begin{figure*}[!htbp]
  \centering
  {\small\bfseries Prefill only, GPU (1024 tokens)\par}
  \includegraphics[width=0.9\textwidth,trim=0 0 0 5bp,clip]{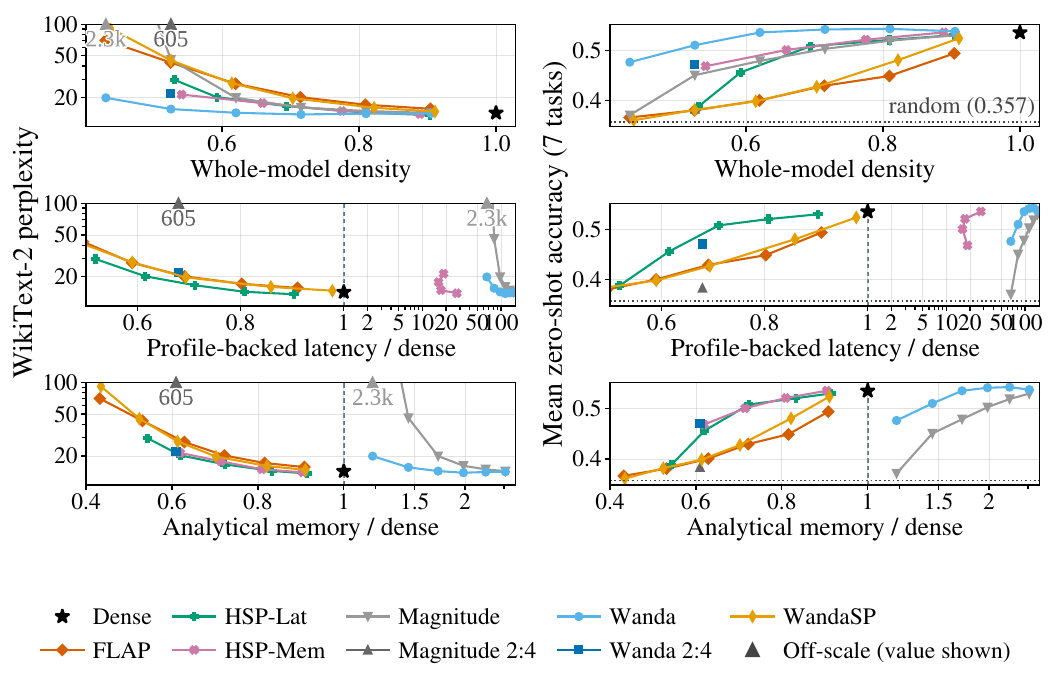}
  \caption{OPT-2.7B GPU prefill results. Plotting conventions follow
    Figure~\ref{fig:frontier_ppl_2p7b_gpu_mixed}.}
  \label{fig:frontier_ppl_2p7b_gpu_prefill}
\end{figure*}

\begin{figure*}[!htbp]
  \centering
  {\small\bfseries Decode only, GPU (128 generated tokens)\par}
  \includegraphics[width=0.9\textwidth,trim=0 0 0 5bp,clip]{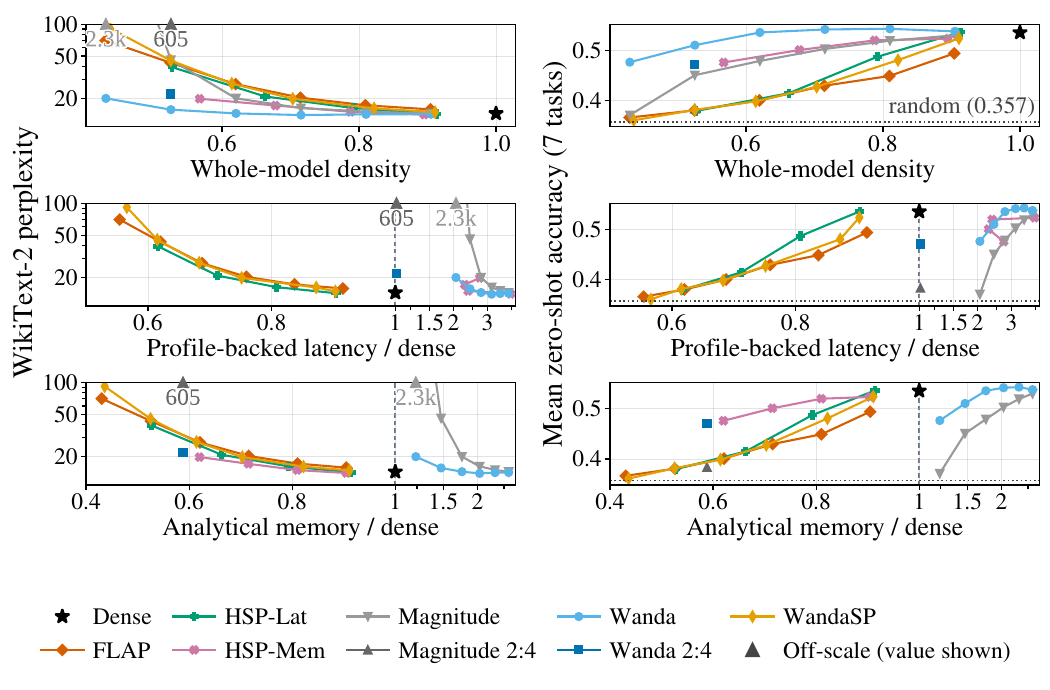}
  \caption{OPT-2.7B GPU decode results. Plotting conventions follow
    Figure~\ref{fig:frontier_ppl_2p7b_gpu_mixed}.}
  \label{fig:frontier_ppl_2p7b_gpu_decode}
\end{figure*}

\begin{figure*}[!htbp]
  \centering
  {\small\bfseries Prefill only, CPU (1024 tokens)\par}
  \includegraphics[width=0.9\textwidth,trim=0 0 0 5bp,clip]{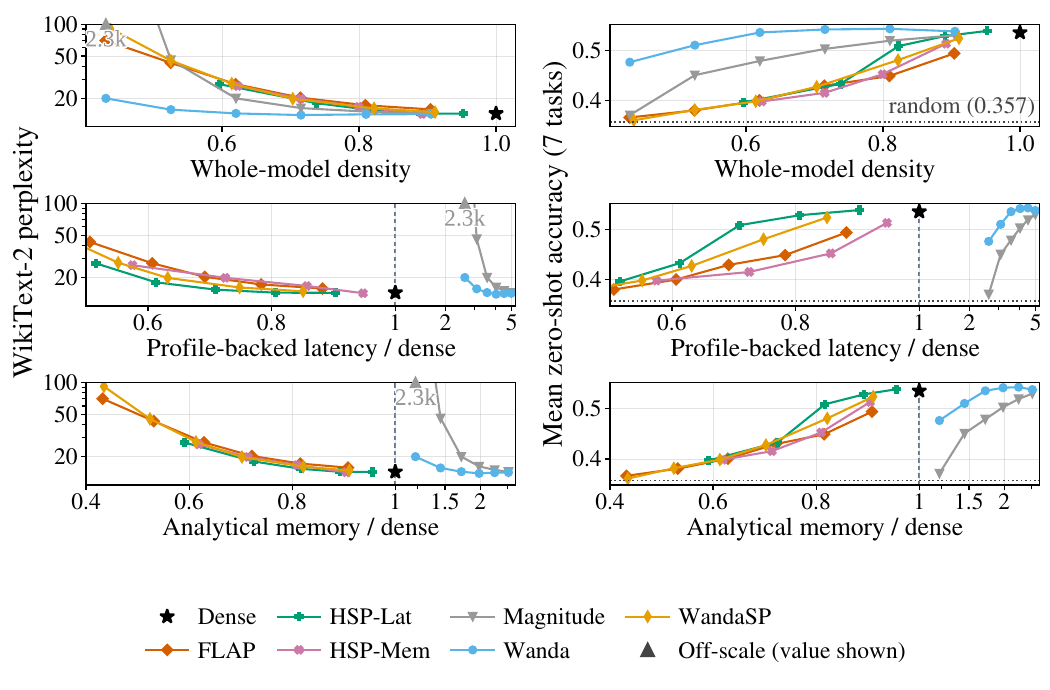}
  \caption{OPT-2.7B CPU prefill results. Plotting conventions follow
    Figure~\ref{fig:frontier_ppl_2p7b_gpu_mixed}.}
  \label{fig:frontier_ppl_2p7b_cpu_prefill}
\end{figure*}

\begin{figure*}[!htbp]
  \centering
  {\small\bfseries Decode only, CPU (128 generated tokens)\par}
  \includegraphics[width=0.9\textwidth,trim=0 0 0 5bp,clip]{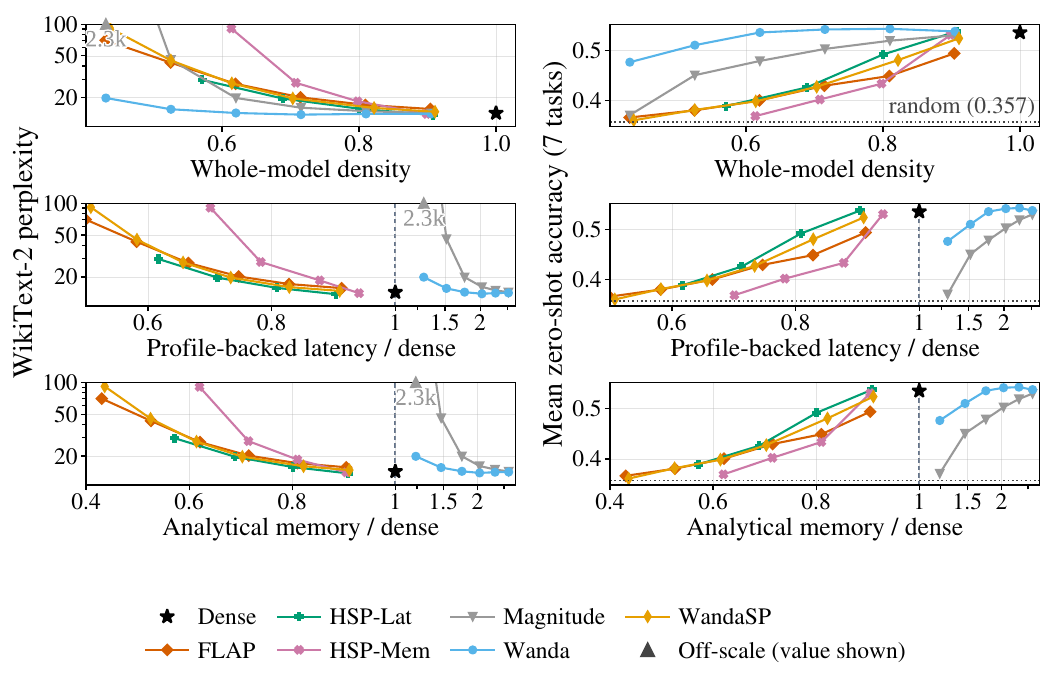}
  \caption{OPT-2.7B CPU decode results. Plotting conventions follow
    Figure~\ref{fig:frontier_ppl_2p7b_gpu_mixed}.}
  \label{fig:frontier_ppl_2p7b_cpu_decode}
\end{figure*}

Figures~\ref{fig:frontier_ppl_2p7b_gpu_prefill}--\ref{fig:frontier_ppl_2p7b_cpu_decode}
compare prefill and decode separately. On the GPU, HSP-Mem preserves quality
at lower memory costs, but its selected structures are much slower for
prefill. HSP-Lat gives a better quality--latency trade-off in this phase.
During decode, the latency curves for HSP-Lat, HSP-Mem, FLAP, and WandaSP
are much closer. On both devices, unstructured pruning preserves quality
at low density but remains slower and uses more memory than dense execution.
The CPU figures omit $2{:}4$ baselines because the runtime has no supported
$2{:}4$ kernel.

\subsection{Scaling Across OPT Models}

\begin{figure*}[!htbp]
  \centering
  {\small\bfseries OPT-125M\par}
  \includegraphics[width=\textwidth]{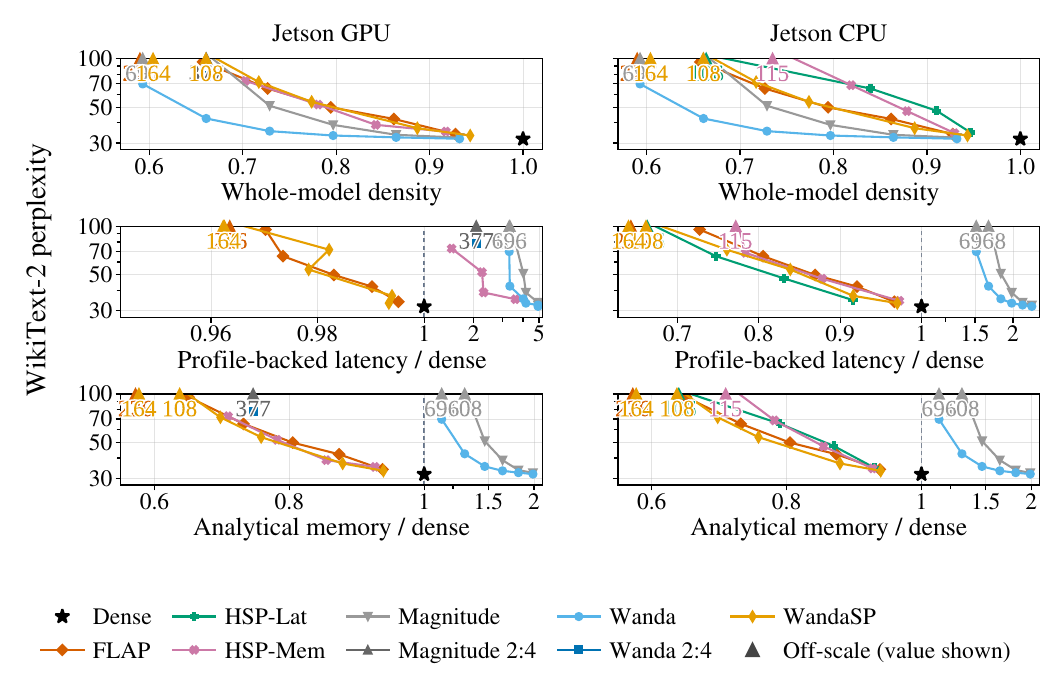}
  \caption{Perplexity versus density, latency, and memory on the GPU (left)
    and CPU (right) under the mixed workload. Plotting conventions follow
    Figure~\ref{fig:frontier_ppl_2p7b_gpu_mixed}.}
  \label{fig:frontier_ppl_125m_gpu_cpu_mixed}
\medskip
  \includegraphics[width=\textwidth]{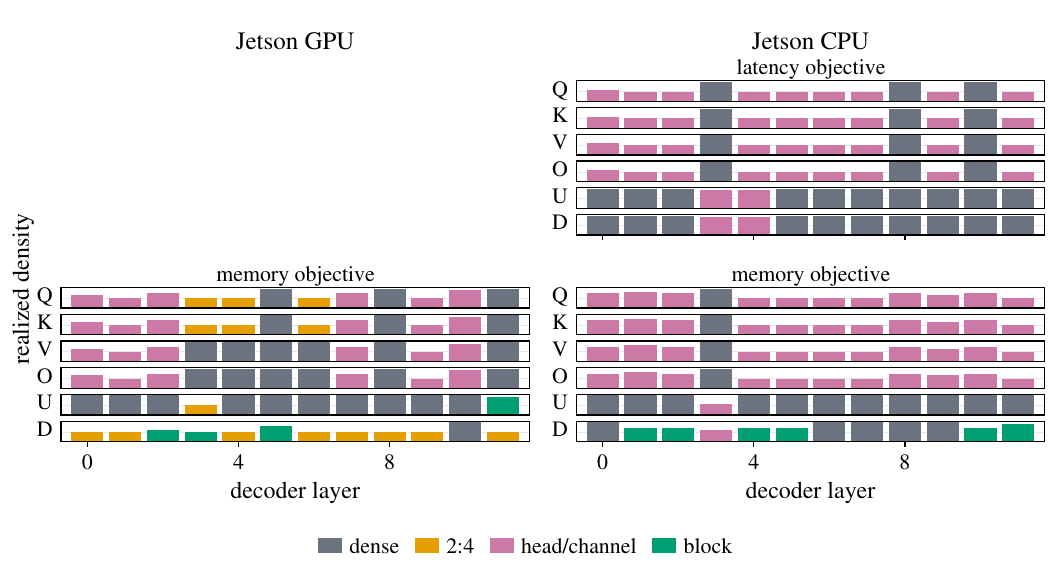}
  \caption{HSP allocations at a budget ratio of $0.8$. Layout and colours
    follow Figure~\ref{fig:structures_0p8_budget}. GPU latency is infeasible.}
  \label{fig:structures_pair_0p8_125m_gpu_cpu}
\end{figure*}

\begin{figure*}[!htbp]
  \centering
  {\small\bfseries OPT-350M\par}
  \includegraphics[width=\textwidth]{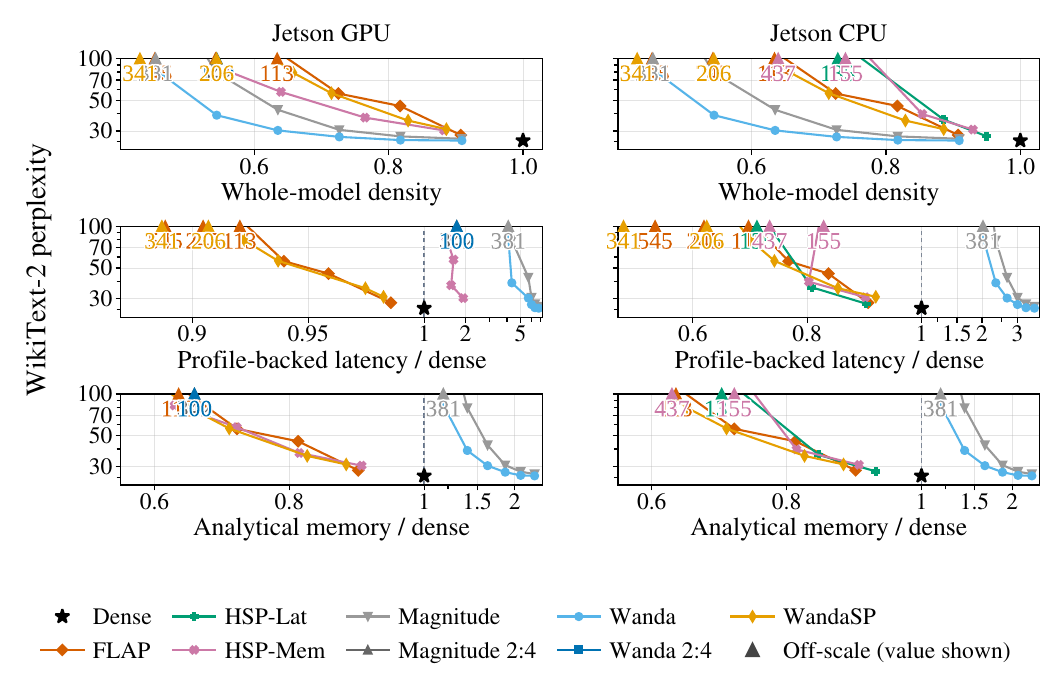}
  \caption{Perplexity versus density, latency, and memory on the GPU (left)
    and CPU (right) under the mixed workload. Plotting conventions follow
    Figure~\ref{fig:frontier_ppl_2p7b_gpu_mixed}.}
  \label{fig:frontier_ppl_350m_gpu_cpu_mixed}
\medskip
  \includegraphics[width=\textwidth]{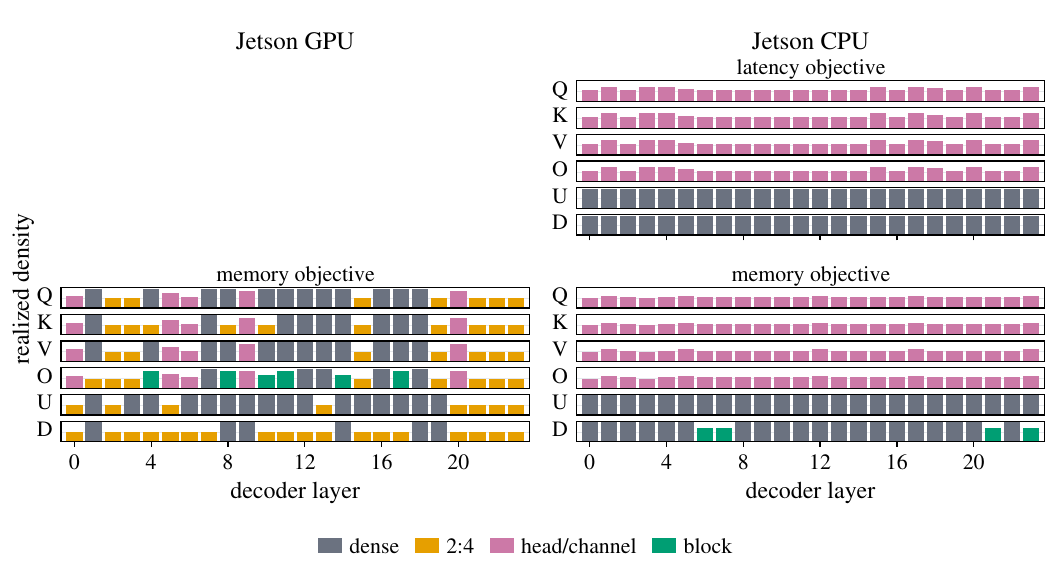}
  \caption{HSP allocations at a budget ratio of $0.8$. Layout and colours
    follow Figure~\ref{fig:structures_0p8_budget}. GPU latency is infeasible.}
  \label{fig:structures_pair_0p8_350m_gpu_cpu}
\end{figure*}

\begin{figure*}[!htbp]
  \centering
  {\small\bfseries OPT-1.3B\par}
  \includegraphics[width=\textwidth]{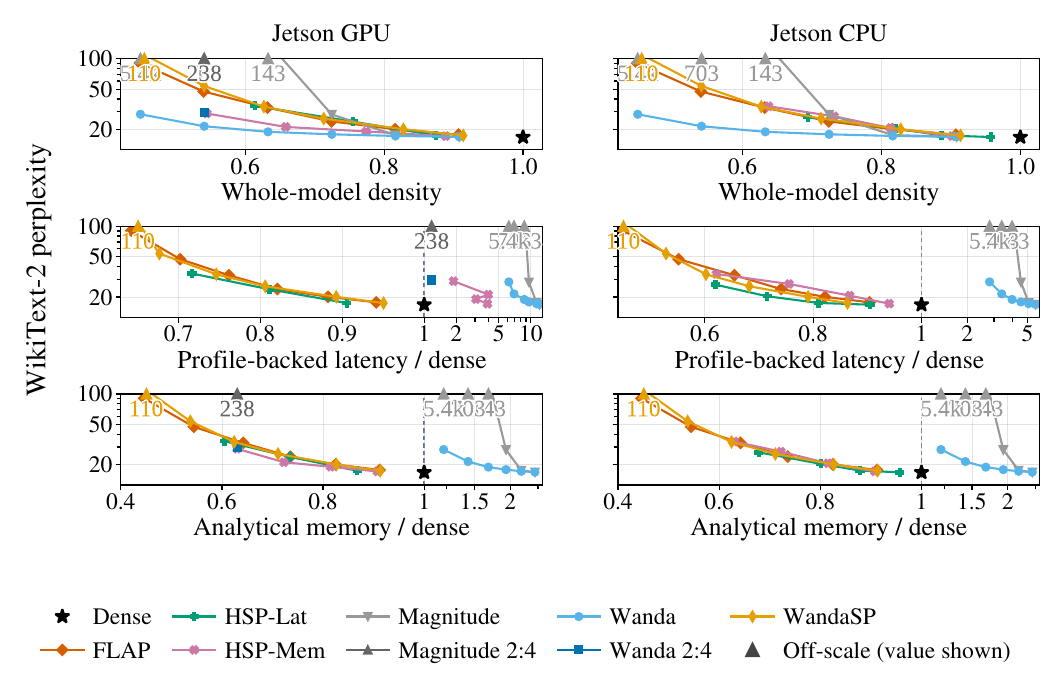}
  \caption{Perplexity versus density, latency, and memory on the GPU (left)
    and CPU (right) under the mixed workload. Plotting conventions follow
    Figure~\ref{fig:frontier_ppl_2p7b_gpu_mixed}.}
  \label{fig:frontier_ppl_1p3b_gpu_cpu_mixed}
\medskip
  \includegraphics[width=\textwidth]{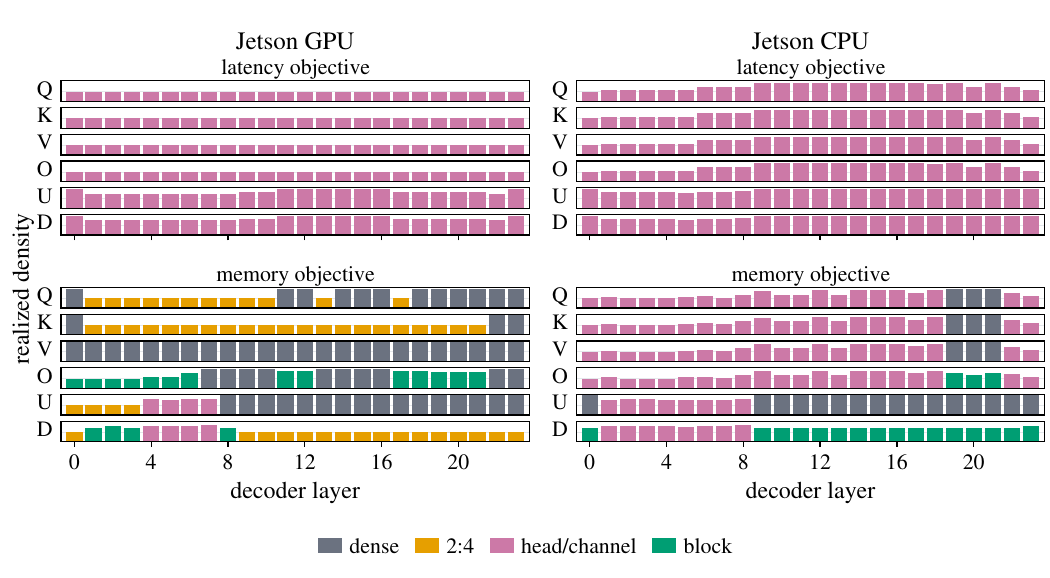}
  \caption{HSP allocations at a budget ratio of $0.8$. Layout and colours
    follow Figure~\ref{fig:structures_0p8_budget}. All four combinations are feasible.}
  \label{fig:structures_pair_0p8_1p3b_gpu_cpu}
\end{figure*}

Figures~\ref{fig:frontier_ppl_125m_gpu_cpu_mixed}--\ref{fig:structures_pair_0p8_1p3b_gpu_cpu}
pair the quality--resource frontiers with the selected structures for
OPT-125M, OPT-350M, and OPT-1.3B. Across these models, lower density does not
reliably translate into lower latency or memory use. HSP mixes structures
and densities across projections and layers, with different allocations for
each model, device, and objective. At a budget ratio of $0.8$, HSP finds
allocations for all combinations except GPU latency for OPT-125M and OPT-350M.

\subsection{LLaMA 7B}

\begin{figure}[p]
  \centering
  {\small\bfseries LLaMA 7B: HSP-Mem\par}\smallskip
  \begin{subfigure}{0.70\linewidth}
    \centering
    \includegraphics[width=\linewidth]{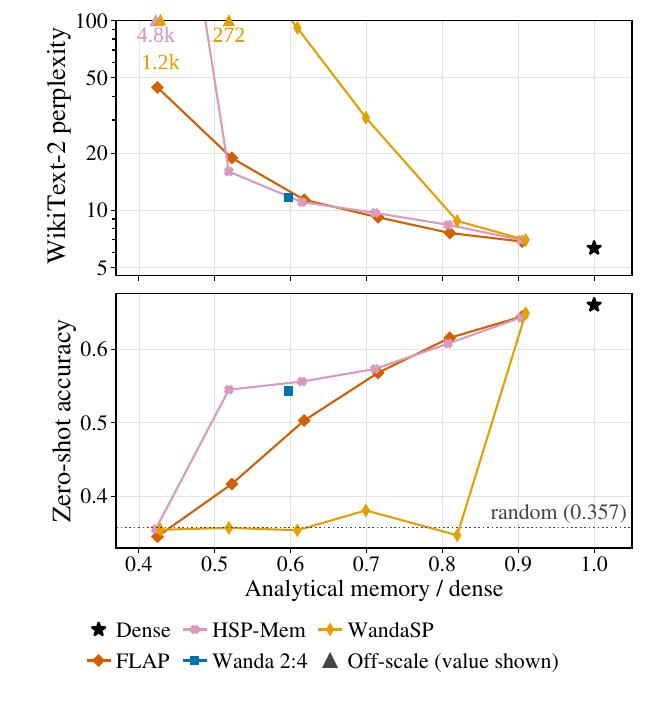}
    \caption{Quality--memory frontiers.}
    \label{fig:app_llama7b_frontiers}
  \end{subfigure}

  \smallskip
  \begin{subfigure}{0.49\linewidth}
    \centering
    \includegraphics[width=\linewidth]{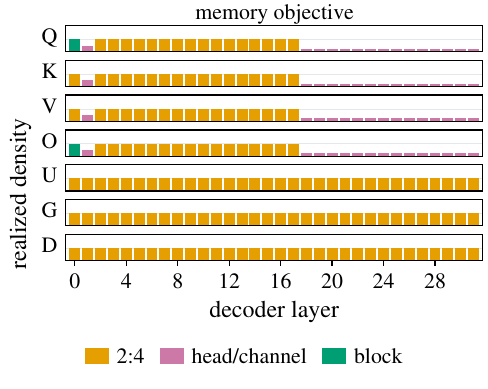}
    \caption{$50\%$ memory constraint.}
    \label{fig:app_llama7b_0p5_structure}
  \end{subfigure}\hfill
  \begin{subfigure}{0.49\linewidth}
    \centering
    \includegraphics[width=\linewidth]{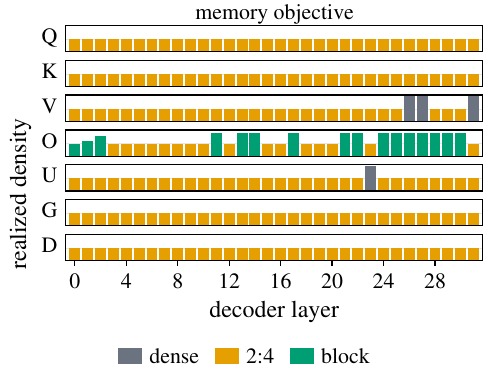}
    \caption{$60\%$ memory constraint.}
    \label{fig:app_llama7b_0p6_structure}
  \end{subfigure}
  \caption{LLaMA 7B with the GPU search space under the mixed workload
    (1024-token prefill, 128-token decode).
    (\subref{fig:app_llama7b_frontiers}) Perplexity and mean seven-task zero-shot
    accuracy versus analytical memory, normalised to dense. Triangles mark
    off-scale perplexity, annotated with its true value. The dotted line marks
    random-guess accuracy (0.357).
    (\subref{fig:app_llama7b_0p5_structure}, \subref{fig:app_llama7b_0p6_structure})
    Structures and retained densities selected for each projection in each
    decoder layer.}
  \label{fig:app_llama7b}
\end{figure}

We evaluate the \texttt{huggyllama/llama-7b} checkpoint and its tokenizer in
FP16 to study whether HSP scales to a larger model from a different family
\citep{touvron2023llamaopenefficientfoundation}.
The FP16 model does not fit in the Jetson Orin Nano's 8 GB of shared
memory even with aggressive pruning, so we evaluate quality offline against analytical memory costs.
We otherwise follow the evaluation setup in Section~\ref{app:experimental_configuration} and widen the GPU search space to include H/C densities as low as $12.5\%$.
Figure~\ref{fig:app_llama7b} reports the resulting frontiers and allocations.
Across the plotted memory range, HSP-Mem preserves perplexity and zero-shot
accuracy more consistently than FLAP~\citep{an2024flap}. It also provides a
variable-memory frontier, whereas fixed-density Wanda-$2{:}4$
\citep{sun2024wanda} offers only a single operating point.

\clearpage

\subsection{Selected Block Shapes}

\begin{figure*}[!htbp]
  \centering
  \includegraphics[width=\textwidth]{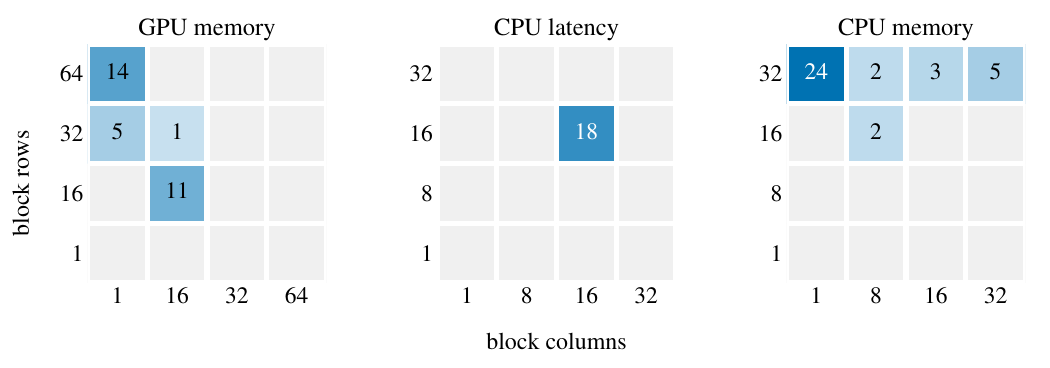}
  \caption{Block shapes selected by the OPT-2.7B candidates in
    Figure~\ref{fig:structures_0p8_budget}. Each cell counts projections
    assigned that shape. Block rows and columns span output and input
    dimensions, respectively. Grey cells indicate zero selections.}
  \label{fig:app_block_shape_histograms}
\end{figure*}

The memory-constrained candidates mostly select narrow blocks, often only
one column wide (Figure~\ref{fig:app_block_shape_histograms}). The CPU latency
candidate instead selects $16\times16$ blocks, while the GPU latency
candidate selects no rectangular blocks.

\begin{samepage}
\subsection{Importance Score and Structure Ablation}
\label{app:importance_ablation}

Wanda gives lower perplexity than magnitude across all workloads and budgets,
with the largest differences under tight constraints
(Table~\ref{table:ablation_latency_variants_ppl}). Without H/C options, HSP cannot meet most of the
decode and mixed latency budgets. Removing these options has a smaller
effect at looser prefill budgets.

\end{samepage}

\begin{table*}[!htbp]
  \centering
  \setlength{\tabcolsep}{6pt}
  {\small
    \begin{tabular}{llcccc}
      \toprule
      Workload & Method & \multicolumn{4}{c}{Latency budget} \\
      \cmidrule(lr){3-6}
      & & 0.6 & 0.7 & 0.8 & 0.9 \\
      \midrule
      Prefill & HSP (Wanda) & \textbf{20.22} & \textbf{16.62} & \textbf{14.41} & \textbf{13.65} \\
      & HSP (Magnitude) & \underline{1.49k} & 539.36 & 47.16 & 15.05 \\
      & HSP (Wanda, no H/C) & -- & \underline{18.26} & \underline{15.07} & \underline{13.69} \\
      \midrule
      Decode & HSP (Wanda) & \textbf{39.21} & \textbf{20.75} & \textbf{16.14} & \textbf{14.15} \\
      & HSP (Magnitude) & \underline{16.58k} & \underline{6.15k} & \underline{5.41k} & 43.61 \\
      & HSP (Wanda, no H/C) & -- & -- & -- & \underline{17.19} \\
      \midrule
      Mixed & HSP (Wanda) & \textbf{38.20} & \textbf{21.28} & \textbf{15.30} & \textbf{14.51} \\
      & HSP (Magnitude) & \underline{16.33k} & \underline{4.86k} & \underline{562.13} & 84.72 \\
      & HSP (Wanda, no H/C) & -- & -- & -- & \underline{16.70} \\
      \bottomrule
    \end{tabular}
  }
  \caption{OPT-2.7B perplexity under latency constraints on the Jetson Orin
    Nano Super GPU. Best and second-best values within each workload and
    budget are bold and underlined, respectively. ``--'' indicates that no
    candidate meets the budget.}
  \label{table:ablation_latency_variants_ppl}
\end{table*}

\subsection{Calibration Sample-Count Sensitivity}
\label{app:calibration}

\begin{table}[H]
  \centering
  \small
  \setlength{\tabcolsep}{3pt}
  \begin{tabular}{ccccc}
    \toprule
    Calibration Samples & 4 & 16 & 64 & 256 \\
    \midrule
    Zero Input-Feature Norms (\%) & 6.44 & 4.83 & 4.52 & 4.46 \\
    Perplexity ($\downarrow$) & 16.947 & 15.129 & 14.880 & 15.299 \\
    \bottomrule
  \end{tabular}
  \caption{Calibration sample-count sensitivity for OPT-2.7B on the GPU
    under the mixed workload, with a latency budget ratio of $0.8$ and all
    other settings fixed. Zero input-feature norms are the percentage of
    projection-input features with zero accumulated $\ell_2$ norm.}
  \label{table:calibration}
\end{table}

Perplexity improves from 4 to 16 calibration samples, with little change
beyond that (Table~\ref{table:calibration}). We use 256 samples throughout
the experiments.

%% file: sections/appendix/notation.tex
\clearpage
\section{Notation Reference}
\label{app:notation_reference}

Superscript $(\pi)$ identifies a projection and $(l)$ a sublayer.
The consumer projection is $\pi_{\mathrm{c}}$ ($O$ or $D$).

\begin{center}
  \small
  \begin{tabular}{@{}p{0.25\columnwidth}@{\hspace{6pt}}p{0.7\columnwidth}@{}}
    \toprule
    Symbol & Meaning \\
    \midrule
    $\pi,\Pi,\Pi_l$ & Projection index, all prunable projections,\newline and projections in sublayer $l$. \\
    $l,L$ & Sublayer index and number of sublayers. \\
    $Q,K,V,O;\ U,G,D$ & Attention projections ($Q,K,V,O$) and MLP projections ($U,G,D$). \\
    $\mathbf{W}^{(\pi)},\mathbf{X}^{(\pi)}$ & Projection weights and calibration-token input activations. \\
    $\mathbf{S}^{(\pi)},\overline{S}^{(\pi)},\widehat{\mathbf{S}}^{(\pi)}$ & Raw importance scores, their mean, and mean-normalised outer scores. \\
    $\mathbf{M}^{(\pi)},\mathcal{M}^{(\pi)}$ & Binary mask and the space of all binary masks of its shape. \\
    $\mathcal{M}^{(\pi)}(s,d)$ & Masks admissible for structure $s$ and density $d$. \\
    $s,d,d_{\min}(s)$ & Sparsity structure, retained density, and structure-specific density floor. \\
    $\phi^{(\pi)},\Phi^{(\pi)}$ & Projection option $(s,d)$ and its finite option set. \\
    $\phi^{(l)},\Phi^{(l)},\boldsymbol{\phi},\Phi$ & Sublayer option and option set, model-wide configuration and configuration space. \\
    $\Phi_{\mathrm{ind}}^{(l)},\Phi_{\mathrm{cpl}}^{(l)},\mathcal{G}_{\mathrm{cpl}}^{(l)}$ & Independent and coupled option sets, and admissible coupled densities. \\
    $\mathbf{M}^{(l)},\mathcal{M}^{(l)}(\phi)$ & Sublayer mask collection and its option-dependent admissible set. \\
    $\ell,\ell_{\mathrm{in}}^{(l)},\ell_{\mathrm{out}}^{(l)}$ & Removed-importance proxy and inner/outer mask losses. \\
    $\mathbf{M}^{\star(l)}(\phi),e_l(\phi)$ & Inner mask solution and its outer loss. \\
    $w^{(l)}(\phi)$ & Positive option weight applied to the outer loss. \\
    $\bar{\mathbf{x}},\mathbf{b},\widetilde{\mathbf{b}}$ & Calibration mean input, original bias, and compensated bias. \\
    $C,C_l,C_0$ & Modelled total, sublayer, and fixed costs. \\
    $C_{\pi},C_{\mathrm{aux}}^{(l)}$ & Projection cost and auxiliary sublayer cost. \\
    $B$ & Absolute latency or memory budget. Sublayers share $B-C_0$. \\
    $D(\mathbf{M}),\rho$ & Aggregate projection-weight density and its budget. \\
    $\mathcal{L},\widetilde{\mathcal{L}}$ & Task loss and aggregate importance-proxy loss. \\
    $\mathcal{D},\mathcal{D}_{\mathrm{cal}}$ & Task-evaluation and calibration datasets. \\
    $\boldsymbol{\phi}^{\star},\Delta,\mathcal{F}$ & Selected configuration, resource-bin width, and dynamic-programming frontier. \\
    $C_{\mathrm{dense}},C_{\mathrm{BSR}},C_{2{:}4}$ & Weight storage in bytes for dense, BSR, and $2{:}4$ formats. \\
    $C_{\mathrm{KV}},T,d_h$ & Cache storage, final cache length, and head dimension. \\
    $H_Q,H_{\mathrm{KV}};\ d_Q,d_{\mathrm{KV}}$ & GQA query/KV head counts and retained fractions. $H_{\mathrm{KV}}^{(k)}$ counts retained heads in the cache model. \\
    $E_{\mathrm{lin}},T_{\mathrm{dense}},T_{\mathrm{sparse}}$ & Linear-scaling efficiency and dense/sparse kernel latencies. \\
    \bottomrule
  \end{tabular}
\end{center}

%% file: refs.bib
@misc{zhang2022opt,
  title         = {OPT: Open Pre-trained Transformer Language Models},
  author        = {Susan Zhang and Stephen Roller and Naman Goyal and Mikel Artetxe and Moya Chen and Shuohui Chen and Christopher Dewan and Mona Diab and Xian Li and Xi Victoria Lin and Todor Mihaylov and Myle Ott and Sam Shleifer and Kurt Shuster and Daniel Simig and Punit Singh Koura and Anjali Sridhar and Tianlu Wang and Luke Zettlemoyer},
  year          = {2022},
  eprint        = {2205.01068},
  archiveprefix = {arXiv},
  primaryclass  = {cs.CL},
  doi           = {10.48550/arXiv.2205.01068},
  url           = {https://arxiv.org/abs/2205.01068}
}

@inproceedings{frantar2023sparsegpt,
  author    = {Elias Frantar and Dan Alistarh},
  title     = {SparseGPT: Massive Language Models Can be Accurately Pruned in One-Shot},
  booktitle = {Proceedings of the International Conference on Machine Learning, {ICML} 2023},
  volume    = {202},
  pages     = {10323--10337},
  publisher = {{PMLR}},
  year      = {2023},
  url       = {https://proceedings.mlr.press/v202/frantar23a.html},
  numpages  = {15}
}

@inproceedings{sun2024wanda,
  author    = {Mingjie Sun and Zhuang Liu and Anna Bair and J. Zico Kolter},
  title     = {A Simple and Effective Pruning Approach for Large Language Models},
  booktitle = {Proceedings of the International Conference on Learning Representations, {ICLR} 2024},
  year      = {2024},
  url       = {https://openreview.net/forum?id=PxoFut3dWW}
}

@inproceedings{an2024flap,
  author    = {An, Yongqi and Zhao, Xu and Yu, Tao and Tang, Ming and Wang, Jinqiao},
  title     = {Fluctuation-based adaptive structured pruning for large language models},
  year      = {2024},
  pages     = {10865--10873},
  isbn      = {978-1-57735-887-9},
  publisher = {AAAI Press},
  url       = {https://doi.org/10.1609/aaai.v38i10.28960},
  doi       = {10.1609/aaai.v38i10.28960},
  booktitle = {Proceedings of the AAAI Conference on Artificial Intelligence, {AAAI} 2024},
  numpages  = {9}
}

@inproceedings{sun2025mdpmultidimensionalvisionmodel,
  title     = {MDP: Multidimensional Vision Model Pruning with Latency Constraint},
  author    = {Xinglong Sun and
               Barath Lakshmanan and
               Maying Shen and
               Shiyi Lan and
               Jingde Chen and
               Jos{\'{e}} M. {\'{A}}lvarez},
  booktitle = {Proceedings of the IEEE/CVF Conference on Computer Vision and Pattern Recognition, {CVPR} 2025},
  pages     = {20113--20123},
  publisher = {Computer Vision Foundation / {IEEE}},
  year      = {2025},
  url       = {https://openaccess.thecvf.com/content/CVPR2025/html/Sun_MDP_Multidimensional_Vision_Model_Pruning_with_Latency_Constraint_CVPR_2025_paper.html},
  doi       = {10.1109/CVPR52734.2025.01873}
}

@inproceedings{shen2026structural_pruning_lat_knapsack,
  author    = {Shen, Maying and Yin, Hongxu and Molchanov, Pavlo and Mao, Lei and Liu, Jianna and Alvarez, Jose M.},
  booktitle = {Advances in Neural Information Processing Systems},
  pages     = {12894--12908},
  publisher = {Curran Associates, Inc.},
  title     = {Structural Pruning via Latency-Saliency Knapsack},
  url       = {https://proceedings.neurips.cc/paper_files/paper/2022/file/5434be94e82c54327bb9dcaf7fca52b6-Paper-Conference.pdf},
  volume    = {35},
  year      = {2022}
}

@inproceedings{han2015magnitudepruning,
  author    = {Han, Song and Pool, Jeff and Tran, John and Dally, William J.},
  title     = {Learning both weights and connections for efficient neural networks},
  year      = {2015},
  publisher = {Curran Associates, Inc.},
  booktitle = {Advances in Neural Information Processing Systems},
  volume    = {28},
  url       = {https://proceedings.neurips.cc/paper/2015/hash/ae0eb3eed39d2bcef4622b2499a05fe6-Abstract.html}
}

@inproceedings{LeCun1989optimal_brain_damage,
  author    = {LeCun, Yann and Denker, John and Solla, Sara},
  booktitle = {Advances in Neural Information Processing Systems},
  editor    = {D. Touretzky},
  publisher = {Morgan-Kaufmann},
  title     = {Optimal Brain Damage},
  url       = {https://proceedings.neurips.cc/paper_files/paper/1989/file/6c9882bbac1c7093bd25041881277658-Paper.pdf},
  volume    = {2},
  pages     = {598--605},
  year      = {1989}
}

@inproceedings{chollet2017xception,
  author    = { Chollet, Francois },
  title     = {{Xception: Deep Learning with Depthwise Separable Convolutions}},
  booktitle = {Proceedings of the IEEE/CVF Conference on Computer Vision and Pattern Recognition, {CVPR} 2017},
  year      = {2017},
  issn      = {1063-6919},
  pages     = {1800-1807},
  doi       = {10.1109/CVPR.2017.195},
  url       = {https://doi.ieeecomputersociety.org/10.1109/CVPR.2017.195},
  publisher = {IEEE Computer Society},
  address   = {Los Alamitos, CA, USA}
}

@inproceedings{Wu2015QuantizedCN,
  title     = {Quantized Convolutional Neural Networks for Mobile Devices},
  author    = {Jiaxiang Wu and Cong Leng and Yuhang Wang and Qinghao Hu and Jian Cheng},
  booktitle = {Proceedings of the IEEE Conference on Computer Vision and Pattern Recognition, {CVPR} 2016},
  year      = {2016},
  pages     = {4820--4828},
  publisher = {IEEE Computer Society},
  address   = {Los Alamitos, CA, USA},
  url       = {https://openaccess.thecvf.com/content_cvpr_2016/html/Wu_Quantized_Convolutional_Neural_CVPR_2016_paper.html}
}

@inproceedings{han2016deepcompression,
  author    = {Song Han and
               Huizi Mao and
               William J. Dally},
  editor    = {Yoshua Bengio and
               Yann LeCun},
  title     = {Deep Compression: Compressing Deep Neural Networks with Pruning, Trained
               Quantization and Huffman Coding},
  booktitle = {Proceedings of the International Conference on Learning Representations, {ICLR} 2016},
  year      = {2016},
  url       = {https://arxiv.org/abs/1510.00149},
  bibsource = {dblp computer science bibliography, https://dblp.org}
}

@article{zhu2024survey_model,
  title     = {A Survey on Model Compression for Large Language Models},
  author    = {Zhu, Xunyu  and
               Li, Jian  and
               Liu, Yong  and
               Ma, Can  and
               Wang, Weiping},
  journal   = {Transactions of the Association for Computational Linguistics, {TACL}},
  volume    = {12},
  year      = {2024},
  address   = {Cambridge, MA},
  publisher = {MIT Press},
  url       = {https://aclanthology.org/2024.tacl-1.85/},
  doi       = {10.1162/tacl_a_00704},
  pages     = {1556--1577}
}

@inproceedings{xia2024shearedllama,
  author    = {Mengzhou Xia and Tianyu Gao and Zhiyuan Zeng and Danqi Chen},
  title     = {Sheared LLaMA: Accelerating Language Model Pre-training via Structured Pruning},
  booktitle = {Proceedings of the International Conference on Learning Representations, {ICLR} 2024},
  year      = {2024},
  url       = {https://openreview.net/forum?id=09iOdaeOzp}
}

@inproceedings{muralidharan2024pruning_and_kd,
  author    = {Muralidharan, Saurav and Turuvekere Sreenivas, Sharath and Joshi, Raviraj and Chochowski, Marcin and Patwary, Mostofa and Shoeybi, Mohammad and Catanzaro, Bryan and Kautz, Jan and Molchanov, Pavlo},
  booktitle = {Advances in Neural Information Processing Systems},
  doi       = {10.52202/079017-1299},
  editor    = {A. Globerson and L. Mackey and D. Belgrave and A. Fan and U. Paquet and J. Tomczak and C. Zhang},
  pages     = {41076--41102},
  publisher = {Curran Associates, Inc.},
  title     = {Compact Language Models via Pruning and Knowledge Distillation},
  url       = {https://proceedings.neurips.cc/paper_files/paper/2024/file/4822991365c962105b1b95b1107d30e5-Paper-Conference.pdf},
  volume    = {37},
  year      = {2024}
}

@inproceedings{men2025shortgpt,
  author    = {Xin Men and Mingyu Xu and Qingyu Zhang and Qianhao Yuan and Bingning Wang and Hongyu Lin and Yaojie Lu and Xianpei Han and Weipeng Chen},
  title     = {ShortGPT: Layers in Large Language Models are More Redundant Than You Expect},
  booktitle = {Findings of the Association for Computational Linguistics, {ACL} 2025},
  pages     = {20192--20204},
  year      = {2025},
  publisher = {Association for Computational Linguistics},
  url       = {https://aclanthology.org/2025.findings-acl.1035/}
}

@inproceedings{song2024sleb,
  author    = {Jiwon Song and Kyungseok Oh and Taesu Kim and Hyungjun Kim and Yulhwa Kim and Jae{-}Joon Kim},
  title     = {{SLEB:} Streamlining LLMs through Redundancy Verification and Elimination of Transformer Blocks},
  booktitle = {Proceedings of the International Conference on Machine Learning, {ICML} 2024},
  publisher = {{PMLR}},
  pages     = {46136--46155},
  year      = {2024},
  volume    = {235},
  numpages  = {20},
  url       = {https://proceedings.mlr.press/v235/song24f.html}
}

@inproceedings{ma2023llm,
  author    = {Xinyin Ma and Gongfan Fang and Xinchao Wang},
  title     = {LLM-Pruner: On the Structural Pruning of Large Language Models},
  booktitle = {Advances in Neural Information Processing Systems},
  volume    = {36},
  pages     = {21702--21720},
  publisher = {Curran Associates, Inc.},
  year      = {2023},
  doi       = {10.52202/075280-0950},
  url       = {https://proceedings.neurips.cc/paper_files/paper/2023/hash/44956951349095f74492a5471128a7e0-Abstract-Conference.html}
}

@inproceedings{ashkboos2024slicegpt,
  author    = {Saleh Ashkboos and Maximilian L. Croci and Marcelo Gennari do Nascimento and Torsten Hoefler and James Hensman},
  title     = {SliceGPT: Compress Large Language Models by Deleting Rows and Columns},
  booktitle = {Proceedings of the International Conference on Learning Representations, {ICLR} 2024},
  year      = {2024},
  url       = {https://openreview.net/forum?id=vXxardq6db}
}

@misc{dalberto2024block,
  title         = {Weight Block Sparsity: Training, Compilation, and AI Engine Accelerators},
  author        = {Paolo D'Alberto and Taehee Jeong and Akshai Jain and Shreyas Manjunath and Mrinal Sarmah and Samuel Hsu and Yaswanth Raparti and Nitesh Pipralia},
  year          = {2024},
  eprint        = {2407.09453},
  archiveprefix = {arXiv},
  primaryclass  = {cs.LG},
  url           = {https://arxiv.org/abs/2407.09453}
}

@misc{wang2026cascadedmultigranularitypruningondevice,
  title         = {Cascaded Multi-Granularity Pruning for On-Device LLM Inference in Industrial IoT},
  author        = {Jinghan Wang and Yanjun Chen and Wei Zhang and Xiaotong Huang and Tianchen Liu and Gaoliang Peng},
  year          = {2026},
  eprint        = {2606.26861},
  archiveprefix = {arXiv},
  primaryclass  = {cs.CL},
  url           = {https://arxiv.org/abs/2606.26861}
}

@inproceedings{kurtic2023ziplm,
  title     = {Zip{LM}: Inference-Aware Structured Pruning of Language Models},
  author    = {Eldar Kurti{\'c} and Elias Frantar and Dan Alistarh},
  booktitle = {Advances in Neural Information Processing Systems},
  volume    = {36},
  pages     = {65597--65617},
  publisher = {Curran Associates, Inc.},
  year      = {2023},
  doi       = {10.52202/075280-2863},
  url       = {https://proceedings.neurips.cc/paper_files/paper/2023/hash/ced46a50befedcb884ccf0cbe8c3ad23-Abstract-Conference.html}
}

@inproceedings{li2021hwnasbench,
  title     = {HW-NAS-Bench: Hardware-Aware Neural Architecture Search Benchmark},
  author    = {Chaojian Li and Zhongzhi Yu and Yonggan Fu and Yongan Zhang and Yang Zhao and Haoran You and Qixuan Yu and Yue Wang and Yingyan (Celine) Lin},
  booktitle = {Proceedings of the International Conference on Learning Representations, {ICLR} 2021},
  year      = {2021},
  url       = {https://openreview.net/forum?id=_0kaDkv3dVf}
}

@article{kjolstad2017taco,
  author     = {Kjolstad, Fredrik and Kamil, Shoaib and Chou, Stephen and Lugato, David and Amarasinghe, Saman},
  title      = {The tensor algebra compiler},
  year       = {2017},
  issue_date = {October 2017},
  publisher  = {Association for Computing Machinery},
  address    = {New York, NY, USA},
  volume     = {1},
  number     = {OOPSLA},
  url        = {https://doi.org/10.1145/3133901},
  doi        = {10.1145/3133901},
  journal    = {Proceedings of the ACM on Programming Languages, {PACMPL}},
  month      = oct,
  eid        = {77},
  numpages   = {29},
}

@misc{gao2024lm_eval,
  author    = {Gao, Leo and Tow, Jonathan and Abbasi, Baber and Biderman, Stella and Black, Sid and DiPofi, Anthony and Foster, Charles and Golding, Laurence and Hsu, Jeffrey and Le Noac'h, Alain and Li, Haonan and McDonell, Kyle and Muennighoff, Niklas and Ociepa, Chris and Phang, Jason and Reynolds, Laria and Schoelkopf, Hailey and Skowron, Aviya and Sutawika, Lintang and Tang, Eric and Thite, Anish and Wang, Ben and Wang, Kevin and Zou, Andy},
  title     = {The Language Model Evaluation Harness},
  month     = jul,
  year      = {2024},
  publisher = {Zenodo},
  version   = {v0.4.3},
  doi       = {10.5281/zenodo.12608602},
  url       = {https://zenodo.org/records/12608602}
}

@inproceedings{clark2019boolq,
  title     = {BoolQ: Exploring the Surprising Difficulty of Natural Yes/No Questions},
  author    = {Christopher Clark and Kenton Lee and Ming-Wei Chang and Tom Kwiatkowski and Michael Collins and Kristina Toutanova},
  booktitle = {Proceedings of the Conference of the North American Chapter of the Association for Computational Linguistics: Human Language Technologies, {NAACL-HLT} 2019},
  year      = {2019},
  pages     = {2924--2936},
  publisher = {Association for Computational Linguistics},
  address   = {Minneapolis, MN, USA},
  url       = {https://doi.org/10.18653/v1/n19-1300},
  doi       = {10.18653/V1/N19-1300}
}

@inproceedings{Bisk2020,
  author    = {Yonatan Bisk and Rowan Zellers and Ronan Le Bras and Jianfeng Gao and Yejin Choi},
  title     = {PIQA: Reasoning about Physical Commonsense in Natural Language},
  booktitle = {Proceedings of the AAAI Conference on Artificial Intelligence, {AAAI} 2020},
  pages     = {7432--7439},
  publisher = {AAAI Press},
  doi       = {10.1609/aaai.v34i05.6239},
  year      = {2020}
}

@inproceedings{zellers2019hellaswag,
  title     = {HellaSwag: Can a Machine Really Finish Your Sentence?},
  author    = {Zellers, Rowan and Holtzman, Ari and Bisk, Yonatan and Farhadi, Ali and Choi, Yejin},
  booktitle = {Proceedings of the Conference of the Association for Computational Linguistics, {ACL} 2019},
  pages     = {4791--4800},
  publisher = {Association for Computational Linguistics},
  address   = {Florence, Italy},
  url       = {https://doi.org/10.18653/v1/p19-1472},
  doi       = {10.18653/V1/P19-1472},
  year      = {2019}
}

@inproceedings{sakaguchi2019winogrande,
  author    = {Keisuke Sakaguchi and
               Ronan Le Bras and
               Chandra Bhagavatula and
               Yejin Choi},
  title     = {WinoGrande: An Adversarial Winograd Schema Challenge at Scale},
  booktitle = {Proceedings of the AAAI Conference on Artificial Intelligence, {AAAI} 2020},
  pages     = {8732--8740},
  publisher = {AAAI Press},
  year      = {2020},
  url       = {https://doi.org/10.1609/aaai.v34i05.6399},
  doi       = {10.1609/AAAI.V34I05.6399},
  bibsource = {dblp computer science bibliography, https://dblp.org}
}

@misc{allenaiarc,
  title         = {Think you have Solved Question Answering? Try ARC, the AI2 Reasoning Challenge},
  author        = {Peter Clark and Isaac Cowhey and Oren Etzioni and Tushar Khot and Ashish Sabharwal and Carissa Schoenick and Oyvind Tafjord},
  year          = {2018},
  eprint        = {1803.05457},
  archiveprefix = {arXiv},
  primaryclass  = {cs.AI},
  url           = {https://arxiv.org/abs/1803.05457}
}

@inproceedings{OpenBookQA2018,
  title     = {Can a Suit of Armor Conduct Electricity? A New Dataset for Open Book Question Answering},
  author    = {Todor Mihaylov and Peter Clark and Tushar Khot and Ashish Sabharwal},
  booktitle = {Proceedings of the Conference on Empirical Methods in Natural Language Processing, {EMNLP} 2018},
  year      = {2018},
  pages     = {2381--2391},
  publisher = {Association for Computational Linguistics},
  doi       = {10.18653/v1/D18-1260},
  url       = {https://doi.org/10.18653/v1/d18-1260}
}

@article{sinha1979mckp,
  issn      = {0030364X, 15265463},
  doi       = {10.1287/opre.27.3.503},
  url       = {http://www.jstor.org/stable/170213},
  author    = {Prabhakant Sinha and Andris A. Zoltners},
  journal   = {Operations Research},
  number    = {3},
  pages     = {503--515},
  publisher = {INFORMS},
  title     = {The Multiple-Choice Knapsack Problem},
  urldate   = {2026-07-28},
  volume    = {27},
  year      = {1979}
}

@book{kellerer2004knapsack,
  author    = {Hans Kellerer and Ulrich Pferschy and David Pisinger},
  title     = {Knapsack Problems},
  publisher = {Springer},
  address   = {Berlin, Heidelberg},
  year      = {2004},
  url       = {https://doi.org/10.1007/978-3-540-24777-7},
  doi       = {10.1007/978-3-540-24777-7}
}

@inproceedings{merity2017pointer,
  title     = {Pointer Sentinel Mixture Models},
  author    = {Stephen Merity and Caiming Xiong and James Bradbury and Richard Socher},
  booktitle = {Proceedings of the International Conference on Learning Representations, {ICLR} 2017},
  year      = {2017},
  url       = {https://openreview.net/forum?id=Byj72udxe}
}

@misc{touvron2023llamaopenefficientfoundation,
  title         = {LLaMA: Open and Efficient Foundation Language Models},
  author        = {Hugo Touvron and Thibaut Lavril and Gautier Izacard and Xavier Martinet and Marie-Anne Lachaux and Timothée Lacroix and Baptiste Rozière and Naman Goyal and Eric Hambro and Faisal Azhar and Aurelien Rodriguez and Armand Joulin and Edouard Grave and Guillaume Lample},
  year          = {2023},
  eprint        = {2302.13971},
  archiveprefix = {arXiv},
  primaryclass  = {cs.CL},
  url           = {https://arxiv.org/abs/2302.13971}
}

@misc{mishra2021sparse,
  title         = {Accelerating Sparse Deep Neural Networks},
  author        = {Asit Mishra and Jorge Albericio Latorre and Jeff Pool and Darko Stosic and Dusan Stosic and Ganesh Venkatesh and Chong Yu and Paulius Micikevicius},
  year          = {2021},
  eprint        = {2104.08378},
  archiveprefix = {arXiv},
  primaryclass  = {cs.LG},
  url           = {https://arxiv.org/abs/2104.08378}
}
